\documentclass{article}
\usepackage{arxiv}
\usepackage[utf8]{inputenc}
\usepackage[T1]{fontenc}
\usepackage{amsthm,amsmath,natbib}
\usepackage{amsfonts,amssymb}
\RequirePackage[colorlinks,citecolor=blue,urlcolor=blue]{hyperref}
\usepackage[toc,page]{appendix}
\usepackage{booktabs}

\usepackage{tikz}
\usetikzlibrary{matrix,chains,positioning,decorations.pathreplacing,arrows}
\usepackage{neuralnetwork}
\usepackage{pgfplots}
\pgfplotsset{compat=1.15}
\usepackage{algorithm}
\usepackage[]{algorithmicx}
\usepackage{algpseudocode}
\usepackage{afterpage}
\usepackage{bbm}
\usepackage{setspace}
\usepackage{comment}
\usepackage[normalem]{ulem}

\newcommand\independent{\protect\mathpalette{\protect\independenT}{\perp}}
\def\independenT#1#2{\mathrel{\rlap{$#1#2$}\mkern2mu{#1#2}}}

\def\qadj{\hat{q}_{\mathrm{adj}}}
\def\tt{\mathbf t}
\def\z{\mathbf z}
\def\R{\mathcal R}
\def\X{\mathbf X}

\def\x{\mathbf x}

\def\r{\mathbf r}

\def\y{\mathbf y}
\def\u{\mathbf u}
\def\v{\mathbf v}

\def\bbeta{\boldsymbol \beta}

\def\t{^\top}

\def \real{\rm I\!R}

\def \r {\mathrm{ReLU}}
\def \P {\mathrm{Pr}}
\def \vahid {\color{magenta}}

\definecolor{dark_green}{rgb}{0.13,0.459,0.133}
\newcommand{\mouloud}{\textcolor{dark_green}}

\renewcommand{\theenumi}{\alph{enumi}}

\newcommand{\norm}[1]{\left\lVert#1\right\rVert}

\usepackage{mathtools}
\DeclarePairedDelimiter{\ceil}{\lceil}{\rceil}
\DeclareMathOperator*{\argmin}{argmin~} 
\DeclareMathOperator{\sign}{sign} 

\newcommand\code{\bgroup\@makeother\_\@makeother\~\@makeother\$\@codex}
\def\@codex#1{{\normalfont\ttfamily\hyphenchar\font=-1 #1}\egroup}
\let\proglang=\textsf
\newcommand{\pkg}[1]{{\fontseries{b}\selectfont #1}}

\newtheorem{theorem}{Theorem}
\newtheorem{definition}{Definition}

\newtheorem{assumption}{Assumption}

\title{A Twin Neural Model for Uplift}

\author{
   Mouloud Belbahri\thanks{Corresponding author: mouloud.belbahri@umontreal.ca} \\
  Department of Mathematics and Statistics\\
  University of Montreal
  \And
  Alejandro Murua\\
  Department of Mathematics and Statistics\\
  University of Montreal
  \And
  Olivier Gandouet\\
  Analytics \& Modeling Advanced Projects \\
  TD Insurance
  \And
  Vahid Partovi Nia \\
  Department of Mathematics and Industrial Engineering\\
  École Polytechnique de Montréal
}

\begin{document}
\maketitle

\begin{abstract}

Uplift is a particular case of conditional treatment effect modeling. Such models deal with cause-and-effect inference for a specific factor, such as a marketing intervention or a medical treatment. In practice, these models are built on individual data from randomized clinical trials where the goal is to partition the participants into heterogeneous groups depending on the uplift. Most existing approaches are adaptations of random forests for the uplift case. Several split criteria have been proposed in the literature, all relying on maximizing heterogeneity. However, in practice, these approaches are prone to overfitting. In this work, we bring a new vision to uplift modeling. We propose a new loss function defined by leveraging a connection with the Bayesian interpretation of the relative risk. Our solution is developed for a specific twin neural network architecture allowing to jointly optimize the marginal probabilities of success for treated and control individuals. We show that this model is a generalization of the uplift logistic interaction model. We modify the stochastic gradient descent algorithm to allow for structured sparse solutions. This helps training our uplift models to a great extent. We show our proposed method is competitive with the state-of-the-art in simulation setting and on real data from large scale randomized experiments.
\end{abstract}

\keywords{causal inference \and heterogeneous treatment effects \and loss function \and gradient descent \and regularization}

\section{Introduction}
\label{sec:intro}

Causal inference draws conclusion about cause and effect relationships through empirical observations. The most widely used statistical framework for causal inference is the \emph{counterfactual} framework. The counterfactual paradigm, also called the \emph{potential outcome} paradigm, was first developed by \cite{Neyman:1923} to study randomized experiments. A generalization allowing the study of causal links with observational data was subsequently carried out by \cite{rubin1974estimating}. Although the process of causal inference is usually complex, it is of extreme importance. In the field of health science, causal inference techniques make it possible to assess the effect of a potential intervention on the health of individuals, in particular in the case of randomized clinical trials. In the field of marketing, such models deal with customers behavioral change caused by a specific treatment, such as a marketing intervention, a courtesy call, targeted advertisement. The counterfactual paradigm assumes that for each individual, there are two potential outcomes, or counterfactuals: i) the potential outcome corresponding to the exposure (treatment), and ii) the potential outcome corresponding to the absence of exposure (control). However, one cannot simultaneously observe treatment and control for a single individual \citep{holland1986statistics}.

In causal inference, the most common population-level estimand is the \emph{average treatment effect}. In the absence of \emph{confounders} (i.e., a variable that influences both the treatment and response variables), this is simply the difference of the two averages between the treatment and control groups. Another estimand include the \emph{individual treatment effect,} also called \emph{conditional average treatment effect} in the context of causal inference, or \emph{uplift} in the context of marketing research. There are different  methods to estimate individual treatment effects. The process of estimating these effects has different synonyms like heterogeneous effect modeling, individual treatment effect modeling, or uplift modeling. 

Uplift modeling is an important research area in marketing field \citep{radcliffe1999differential, hansotia2002direct, lo2002true, radcliffe2007using}. 
This modeling framework has also been proposed to allow for prediction of an individual patient’s response to a medical treatment \citep{jaskowski2012uplift,lamont2018identification}.
Typically, uplift models are developed for randomized experiments, with both the treatment and outcome as binary random variables, where prediction power is the most important issue. 

Most common research in the uplift field is based on classification and regression trees (CART) \citep{breiman1984classification}. Unlike other modeling techniques, fitting a decision tree allows each iteration to uniquely partition the sample. This means that each segmentation can be immediately checked against the impact of the treatment. Since the goal of uplift modeling is to find a partition into subgroups of the population, it seems natural to use a decision tree as the method of choice. Then, the idea is to predict the individual uplift by the uplift observed in a terminal node, or by the average when several trees are used, e.g., random forests \citep{Breiman-RandomForest-2001}. However, this is not enough. Indeed, with CART classic division criteria, the search for predictors and their split points is optimized according to the response variable and not the uplift. As practice often shows, response probability and uplift have quite different factors in terms of predictors \citep{radcliffe2011real}. In extreme cases, all uplifts in the leaves might be the same (but not the positive response probabilities). Therefore, various modifications of the split criteria have been proposed in the literature \citep{radcliffe1999differential, hansotia2002direct, rzepakowski2010decision}.

In the case of parametric modeling approaches, the simplest model that can be used to estimate the uplift is logistic regression, because the response variable is binary \citep{lo2002true}. Thus, the underlying optimization problem becomes the maximization of the Binomial likelihood. In this case, the approach does not provide a direct uplift search, but rather the probabilities of positive responses for treated and non-treated are modeled separately. Then, their difference is used as an estimate of the uplift.
However, the solution is not optimized for searching for heterogeneous groups depending on the uplift. Hence, maximizing the likelihood is not necessarily the right way to estimate the uplift. Therefore, changes are required in the optimization problem in order to appropriately estimate the uplift. 

In this work, we introduce a new uplift loss function defined by leveraging a connection with the Bayesian interpretation of the relative risk, another treatment effect measure specific to the binary case. Defining an appropriate loss function for uplift also allows to use simple models or any off the shelf method for the related optimization problem, including complex models such as neural networks. When prediction becomes more important than estimation, neural networks become more attractive than classical statistical models. There are several reasons why neural networks are suitable tools for uplift: i) they are flexible models and easy to train with current GPU hardware; ii) with a single hidden layer modeling covariates interactions is straightforward \citep{tsang2018neural}; iii) they are guaranteed to approximate a large class of  functions \citep{cybenko1989approximation, hornik1991approximation, pinkus1999approximation}; iv) neural networks perform very well on predictive tasks which is the main objective of uplift; v) a simple network architecture ensures model interpretability for further studies.

Our methodology is developed in the context of a specific neural network architecture in which the proposed loss function can be easily optimized. Our solution uses a representation that resembles the twin networks \citep{bromley1994signature} known in the context of deep learning. This representation helps the fitting process to a great extent. We show our model generalizes Lo's uplift logistic interaction model to a $1$-hidden layer neural network model. In the logistic regression context, one can use \emph{lasso} \citep{Tibshirani_lasso_1996} to produce a sparse model. As more hidden nodes are added to the neural network, lasso becomes choosing the right number of parameters to include in the network (i.e., pruning). However, the fitting mechanism of statistically-driven methods such as the lasso needs to be adapted for stochastic gradient descent to provide a neural network pruning technique. 
We propose to use a proximal version of gradient descent and to introduce a scaling factor for structured pruning, which is common in neural networks \citep{ramakrishnan2020differentiable}. This allows to select automatically the number of neurons for each hidden layer. Compared to other existing methods, we thus offer a unified principled approach to obtain a sparse solution that provides well-optimized uplift estimates.

The contributions of this paper are the following: i) defining a new loss function derived from an intuitive interpretation of treatment effects estimation; ii) generalizing the uplift logistic interaction model to a $1$-hidden layer ReLU neural-network; iii) introducing a new twin neural architecture to predict conditional average treatment effects;  iv) guiding model selection (or architecture search) with sparse group lasso regularization; v) establishing empirically the validity of the estimation procedure on both synthetic and real-world datasets. 

\section{Related work}
\label{sec:notation}

Uplift is a particular case of conditional treatment effect modeling which falls within the potential outcomes framework, also known as the Neyman-Rubin causal model \citep{rubin1974estimating,rosenbaum1983central,holland1986statistics}.

Let $T$ be the binary  treatment indicator, and $\boldsymbol X = (X_1,\ldots,X_p)$ be the $p$-dimensional predictors vector. The binary variable $T$ indicates if a unit is exposed to treatment ($T=1$) or control ($T=0$). Let $Y(0)$  and $Y(1)$ be the binary potential outcomes under control and treatment respectively. Assume a distribution $(Y(0),Y(1),\boldsymbol X,T) \sim \mathcal{P}$ from which $n$ iid samples are given as the training observations $\{(y_i, \textbf{x}_i, t_i)\}_{i=1}^n$, where $\x_i = (x_{i1}, \ldots, x_{ip})$ are realisations of the predictors and $t_i$ the realisation of the treatment for observation $i$. Although each observation $i$ is associated with two potential outcomes, only one of them can be realized as the observed outcome $y_i$. By Assumption~\ref{ass:assumption3}, under the counterfactual consistency, each observation is missing only one potential outcome: the one that corresponds to the absent treatment either $t=0$ or $t=1$.\\

\begin{assumption}
(Consistency) Observed outcome $Y$ is represented using the potential outcomes and treatment assignment indicator as follows:
$$Y=TY(1) + (1-T)Y(0).$$
\label{ass:assumption3}
\end{assumption}

In general, we will assume the following representation of $\mathcal{P}$:
\begin{align*}
 &\boldsymbol X \sim \Lambda\\
 &T \sim \mathrm{Bernoulli}(e(\x))\\
 &Y(t) \sim \mathrm{Bernoulli}(m_{1t}(\x))
\end{align*}
where $\Lambda$ is the marginal distribution of $\boldsymbol X$ and $e(\cdot)$ is the propensity score (see Definition~\ref{def:propensity_score}). The probabilities of positive responses for the potential outcomes under control and treatment are given by the functions $m_{1t}(\cdot):\real^{p}$ $\rightarrow$ $(0,1)$ for $t=0$ and $t=1$ respectively.

\begin{definition}
(Propensity score) For any $\boldsymbol X=\x$, the \textit{propensity score} is defined as: 
\begin{align}
    e(\x)= \mathrm{Pr}(T_i=1 \mid \boldsymbol X_i = \x).
    \label{def:propensity_score}
\end{align}
\end{definition}

Given the notation above, the conditional average treatment effect (CATE) is defined as follows:
\begin{align}
    \mathrm{CATE}(\x) = \mathbb{E}[Y_i(1) - Y_i(0)| \boldsymbol X_i = \x]
    \label{def:cate}
\end{align}

In order for the CATE to be identifiable, we must make some additional assumptions, standard in the world of causal inference. The propensity score parameter is widely used to estimate treatment effects from observational data \citep{rosenbaum1983central}. Assumption \ref{ass:assumption2} states that each individual has non-zero probabilities of being exposed and being unexposed. This is necessary to make the mean quantities meaningful.

\begin{assumption}
(Overlap) For any $\boldsymbol X=\x$, the true propensity score is strictly between $0$ and $1$, i.e., for $\epsilon > 0$,
$$\epsilon < e(\x) < 1-\epsilon.$$
\label{ass:assumption2}
\end{assumption} 

For the rest of the paper, we will consider the case of randomized experiments, with $e(\x)=1/2$, which is common in the uplift literature and is the case of our data. When the data comes from observational studies, combined with the previous assumptions, Assumption \ref{ass:assumption1} allows the identification of the CATE.
\begin{assumption}
(Unconfoundedness) Potential outcomes $Y(0),Y(1)$ are independent ($\independent$) of the treatment assignment indicator $T$ conditioned on all pre-treatment characteristics $\boldsymbol X$, i.e.,
$$Y(0), Y(1)  \independent T | \boldsymbol X. $$
\label{ass:assumption1}
\end{assumption}

For randomized experiments, the random variable $ T $ is independent of any pre-treatment characteristics, that is, $Y(0), Y(1), \boldsymbol X  \independent T$, which is a stronger assumption than the unconfoundedness assumption.

In the literature, the terms ITE, CATE and uplift often refer to the same quantity, namely the CATE. Indeed, the uplift is a special case of the CATE when the dependent variable $ Y $ is binary $ 0-1 $. Thus, the uplift is defined as the conditional average treatment effect in different sub-populations according to the possible values of the covariates, namely:
\begin{equation}
    u(\x) = \mathrm{Pr}(Y_i=1 \mid \boldsymbol X_i=\x, T_i=1) - \mathrm{Pr}(Y_i=1 \mid \boldsymbol X_i=\x, T_i=0).
    \label{def:uplift}
\end{equation}

To simplify the notation, we prefer to denote by $m_{yt}(\x)$ the corresponding conditional probability $\mathrm{Pr}(Y_i=y \mid \boldsymbol X_i=\x, T_i=t)$. Therefore, the uplift is the difference between the two conditional means $m_{11}(\x)$ and $m_{10}(\x)$. The intuitive approach to model uplift is to build two independent models \citep{hansotia2002direct}. This consists of fitting two separate conditional probability models: one model for the treated individuals, and another separate model for the untreated individuals. Then, uplift is the difference between these two conditional probability models. These models are called T-learners (T for ``two models") in the literature \citep{kunzel2019metalearners}. The asset of T-learners is their simplicity, but they do not perform well in practice, because each model focuses on predicting only  one class, so the information about the other treatment is never provided to the learning algorithm \citep{radcliffe2011real}. In addition, differences between the covariates distributions in the two treatment groups can lead to bias in treatment effect estimation. There has been efforts in correcting such drawbacks through a combined classification model known as S-learner, for ``single-model" \citep{kunzel2019metalearners}. The idea behind the S-learner is to use the treatment variable as a feature and to add explicit interaction terms between each covariate and the treatment indicator to  fit a model, e.g., a logistic regression \citep{lo2002true}. The parameters of the interaction terms measure the additional effect of each covariate due to treatment.

Another related method is known as the X-learner \citep{kunzel2019metalearners}. The X-learner estimates the uplift in three stages. First, $m_{11}(\x)$ and $m_{10}(\x)$ are modeled separately as in the case of T-learners. Then, the fitted values $\hat m_{11}(\x)$ and $\hat m_{10}(\x)$ are used to impute the ``missing" potential outcomes for each observation, and to create new imputed response variables, $D(1)$ and $D(0)$. These imputed variables are used to fit new models that capture the uplift directly, $\hat u^{(1)}(\x)$ and $\hat u^{(0)}(\x)$. The final prediction is given by a weighted average using the propensity score, $e(\x)\{\hat u^{(1)}(\x) - \hat u^{(0)}(\x)\} + \hat u^{(0)}(\x)$. More recently, \cite{NieWager2020Quasi} introduced the R-learner. The method uses Robinson's transformation \citep{robinson1988root} and assumes ``oracle" estimation of the propensity score $e(\x)$ and the marginal effect function $m(\x) = \mathbb{E}[Y|\boldsymbol X=\x]$ in order to reduce the problem of modeling the uplift to a residual-on-residual ordinary least squares regression.  Interestingly, even though the X-learner and R-learner frameworks are valid for continuous and binary $ Y $, both methods aim to find an uplift estimator that minimizes the mean squared error (MSE). The connection with the maximum likelihood  framework does not hold in the binary case. Also, in practice, the R-learner is fitted in two stages using cross-fitting to emulate the oracle, thus requiring more observations than S-learners. 
The methodology introduced in \cite{belba2019qbased} attacks parameters estimation and addresses the loss-metric mismatch in uplift regression. It can be seen as an S-learner which estimates the model's parameters in two stages. The method first applies the pathwise coordinate descent algorithm \citep{friedman2007pathwise} to compute a sequence of critical regularization values and corresponding (sparse) model parameters. Then, it uses Latin hypercube sampling \citep{mckay2000comparison} to explore the parameters space in order to find the optimal model.

Several proposed non-parametric methods take advantage of grouped observations in order to model the uplift directly. Some $k$-nearest neighbours \citep{cover1967nearest} based methods are adopted for uplift estimation \citep{crump2008nonparametric, Alemi.etal-PersonalizedMedicine-2009, su2012facilitating}. The main idea is to estimate the uplift for an observation based on its neighbourhood containing at least one treated and one control observations. However, these methods quickly become computationally expensive for large datasets.
State-of-the-art proposed methods view random forests as an adaptive neighborhood metric, and estimate the treatment effect at the leaf node \citep{su2009subgroup,chipman2010bart,wager2018estimation}. Therefore, most active research in uplift modeling is in the direction of classification and regression trees \citep{breiman1984classification} where the majority are modified random forests \citep{Breiman-RandomForest-2001}. 
In \cite{radcliffe1999differential, radcliffe2011real,hansotia2002direct, rzepakowski2010decision}, modified split criteria that suited the uplift purpose were studied. The criteria used for choosing each split during the growth of the uplift trees is based on maximization of the difference in uplifts between the two child nodes. Within each leaf, uplift is estimated with the difference between the two conditional means. A good estimate of each mean may lead to a poor estimate of the difference \citep{radcliffe2011real}. However, the existing tree-based uplift optimization problems do not take this common misconception into account. Instead, the focus is on maximizing the heterogeneity in treatment effects. Without careful regularization (e.g., honest estimation \citep{athey2019generalized}), splits are likely to be placed next to extreme values because outliers of any treatment can influence the choice of a split point. In addition, successive splits tend to group together similar extreme values, introducing more variance in the prediction of uplift \citep{zhao2017practically}. Alternatively some models use the transformed outcome \citep{athey2015machine}, an unbiased estimator of the uplift. However, this estimate suffers from higher variance than the difference in conditional means estimator \citep{powers2018some}. In addition, for both estimators, if random noise is larger than the treatment effect, the model will more likely predict random noise instead of uplift. As a result, based on several experiments on real data, and although the literature suggests that tree-based methods are state-of-the-art for uplift \citep{soltys2015ensemble}, the published models overfit the training data and predicting uplift still lacks satisfactory solutions. 






\section{An uplift loss function}

We formally define the uplift loss function that will be used to fit our models. Our goal is to regularize the conditional means in order to get a better prediction of the quantity of interest, the uplift. Inspired by the work of \cite{athey2019generalized,kunzel2019metalearners,belba2019qbased,NieWager2020Quasi} which adapt the optimization problem to the uplift context, we propose to define a composite loss function, which can be separated into two pieces: $$\ell(\cdot) = \ell_1(\cdot) + \ell_2(\cdot)$$ and to optimize both simultaneously. Since we generalize the uplift logistic regression, we model the probability of positive response $m_{1t}(\x)$. Naturally, the first term can be defined as the negative log-likelihood or the binary cross entropy (BCE) loss, with $\y$ as the response, and $m_{1t}(\x)$ as the prediction, that is, $$ \ell_1(\y, \tt \mid \x) = - \frac{1}{n} \sum_{i=1}^n \Big(y_i \log\{ m_{ 1 t_i}(\x_i)\} + (1 - y_i)\log\{1 - m_{1 t_i }(\x_i)\} \Big).$$

We define the second term based on a Bayesian interpretation of another measure of treatment effect, the relative risk. First, let us define the relative risk (or risk ratio) as a function of the conditional means.\\

\begin{definition}
(Relative risk) For any $\boldsymbol X = \x$ and $m_{10}(\x) > 0$, the relative risk is defined as follows:
\begin{align*}
    \mathrm{RR}(\x) = \frac{\mathrm{Pr}(Y=1 \mid \boldsymbol X = \x, T=1)}{\mathrm{Pr}(Y=1 \mid \boldsymbol X = \x, T=0)} = \frac{m_{11}(\x)}{m_{10}(\x)},
\end{align*}
\end{definition}
Relative risk is commonly used to present the results of randomized controlled trials. In the medical context, the uplift is known as the absolute risk (or risk difference). In practice, presentation of both absolute and relative measures is recommended \citep{moher2012consort}. If the relative risk is presented without the absolute measure, in cases where the base rate of the outcome $m_{10}(\x)$ is low, large or small values of relative risk may not translate to significant effects, and the importance of the effects to the public health can be overestimated. Equivalently, in cases where the base rate of the outcome $m_{10}(\x)$ is high, values of the relative risk close to 1 may still result in a significant effect, and their effects can be underestimated. Interestingly, the relative risk can be reformulated as:
\begin{align*}
    \mathrm{RR}(\x) = \frac{\mathrm{Pr}(T=1 \mid Y=1, \boldsymbol X = \x)}{\mathrm{Pr}(T=0 \mid Y=1, \boldsymbol X = \x)} \biggl(\frac{1-e(\x)}{e(\x)}\biggr),
\end{align*}
where the propensity score $e(\x)$ is given in Definition~\ref{def:propensity_score}. For randomized experiments, the propensity score ratio $\{1-e(\x)\}/e(\x)$ is a constant and, written in that form, the relative risk can be interpreted in Bayesian terms as the normalized posterior propensity score ratio (i.e., after observing the outcome). In the particular case where $e(\x)=1/2$, it is easy to show that $\mathrm{Pr}(T=1 \mid Y=1, \boldsymbol X = \x) = \mathrm{RR}(\x)/\{1+\mathrm{RR}(\x)\}$. Moreover, we have the following equalities:
\begin{align}
     \mathrm{Pr}(T=1 \mid Y=1, \boldsymbol X = \x) &= \frac{m_{11}(\x)}{m_{11}(\x)+m_{10}(\x)}, \\
     \mathrm{Pr}(T=1 \mid Y=0, \boldsymbol X = \x) &= \frac{m_{01}(\x)}{m_{01}(\x) + m_{00}(\x)}.
\end{align}

These two equalities give a lot of information. Without abuse of language, we call them \textit{posterior propensity scores} and we denote by $p_{yt} (\x)$ the corresponding conditional probability $\mathrm{Pr}(T=t \mid Y=y, \boldsymbol X = \x)$. The posterior propensity scores are functions of the conditional means. 
The quantity $p_{11}(\x)$ can be seen as the proportion of treated observations among those that had positive outcomes and $p_{01}(\x)$ can be seen as the proportion of treated observations among those that had negative outcomes. %
We define the second term of our uplift loss as the BCE loss, but this time, using the observed treatment indicator $\tt$ as the ``response" variable, and $p_{y1}(\x)$ as the ``prediction". Formally, it is given by: $$ \ell_2(\tt, \y \mid \x) = -\frac{1}{n} \sum_{i=1}^n \Big(t_i \log\{p_{y_i 1}(\x_i)\} + (1 - t_i)\log\{1 - p_{y_i 1}(\x_i)\} \Big).$$

Taken alone, the second loss models the posterior propensity scores as a function of the conditional means (for positive and negative outcomes). Intuitively, if a treatment has a significant positive (resp. negative) effect on a sub-sample of observations, then within the sample of observations that had a positive (resp. negative) response, we expect a higher proportion of treated. Formally, we define the complete uplift loss function in Definition~\ref{def:augmented_loss}.\\

\begin{definition}\label{def:augmented_loss}
Let $m_{yt} \stackrel{\text{def}}{=} m_{yt}(\x) = \mathrm{Pr}( Y=y |\boldsymbol X =\x, T=t)$, and $p_{yt} \stackrel{\text{def}}{=} p_{yt} (\x) = m_{yt} / ( m_{y1} + m_{y0}) $. We define the uplift loss function as follows:
\begin{equation}
    \ell(\y , \tt \mid \x) = - \frac{1}{n} \sum_{i=1}^n \{y_i \log m_{ 1 t_i} + (1 - y_i)\log m_{0t_i } + t_i \log p_{y_i 1} + (1 - t_i)\log p_{y_i 0} \} .
    \label{eq:augmented_loss_function}
\end{equation}
\end{definition}

Although we considered the case where $e(\x)=1/2$, the development holds for any constant $e(\x)$. An under-sampling or an over-sampling procedure allows to recover the $e(\x) = 1/2$ if the constant is below or above $1/2$ respectively. Interestingly, it is possible to find a connection between the uplift loss function \eqref{eq:augmented_loss_function} and the likelihood of the data. Indeed, the described relation between the relative risk and the conditional probabilities $m_{11}(\x)$ and $m_{10}(\x)$, as well as the Bayesian interpretation of the posterior propensity scores $p_{yt}(\x)$ suggest modeling the joint distribution of $Y$ and $T$. Formally, the connection can be shown through the following development.
\begin{align*}
    \P(Y=y, T=t \mid \X=\x)  &= \P(T=t \mid Y=y, \X=\x) \P(Y=y\mid \X=\x) \\
    & = p_{yt}(\x) \{ m_{y1}(\x) e(\x) + m_{y0}(\x) [1-e(\x)]\} \\
    & = p_{yt}(\x) \{ m_{y1}(\x) + m_{y0}(\x)\} / 2,
\end{align*}
because $e(\x)=1/2$. Therefore, the likelihood for $n$ observations is proportional to
\begin{align*}
    \prod_{i=1}^n p_{y_i 1}^{t_i} p_{y_i 0}^{(1-t_i)} \{m_{11} + m_{10}\}^{y_i}  \{m_{01} + m_{00}\}^{(1-y_i)},
\end{align*}
and the log-likelihood is proportional to
\begin{equation}
   \sum_{i=1}^n \{y_i \log(m_{11}+m_{10}) + (1 - y_i)\log(m_{01} + m_{00}) + t_i \log p_{y_i 1} + (1 - t_i)\log p_{y_i 0} \}.
    \label{eq:log-likelihood_loss_function}
\end{equation}
Notice that the functions \eqref{eq:augmented_loss_function} and \eqref{eq:log-likelihood_loss_function} differ only in that \eqref{eq:augmented_loss_function} uses $m_{1t}$ while \eqref{eq:log-likelihood_loss_function} specifically uses $m_{y1}$ and $m_{y0}$, the conditional means under treatment and control. Traditionally, $m_{1t}$ is more common since in practice, each observation can only be treated or not treated. However, we compared the results by fitting uplift models using both functions. The results being very similar, in the rest of the paper, we only present results for models fitted with the augmented loss function \eqref{eq:augmented_loss_function}. We keep the in-depth analysis of the log-likelihood \eqref{eq:log-likelihood_loss_function} for future work.

The loss function \eqref{eq:augmented_loss_function} can also be interpreted term by term. The first term is simply the binary cross entropy loss w.r.t the conditional means. The second term can be seen as a regularization term on the conditional means. In the second term, the conditional means are represented through the posterior propensity scores. By minimizing the augmented loss, the first term focuses on estimating the conditional means separately while the second term tries to correct for the posterior propensity scores. Since both terms are minimized simultaneously, this can also be seen as a special case of multi-task learning. As we will show later, this new parameter estimation method greatly improves the predictive performance of the underlying uplift models.

\section{A twin neural model for uplift}

Let's start with the uplift interaction model \citep{lo2002true} as a simple preliminary model. This model is based on logistic regression. It is common to add explicit interaction terms between each explanatory variable and the treatment indicator. The parameters of the interaction terms measure the additional effect of each covariate due to treatment. These interactions are important for estimating individual effects since this is what makes it possible to create heterogeneity in the treatment effects. 

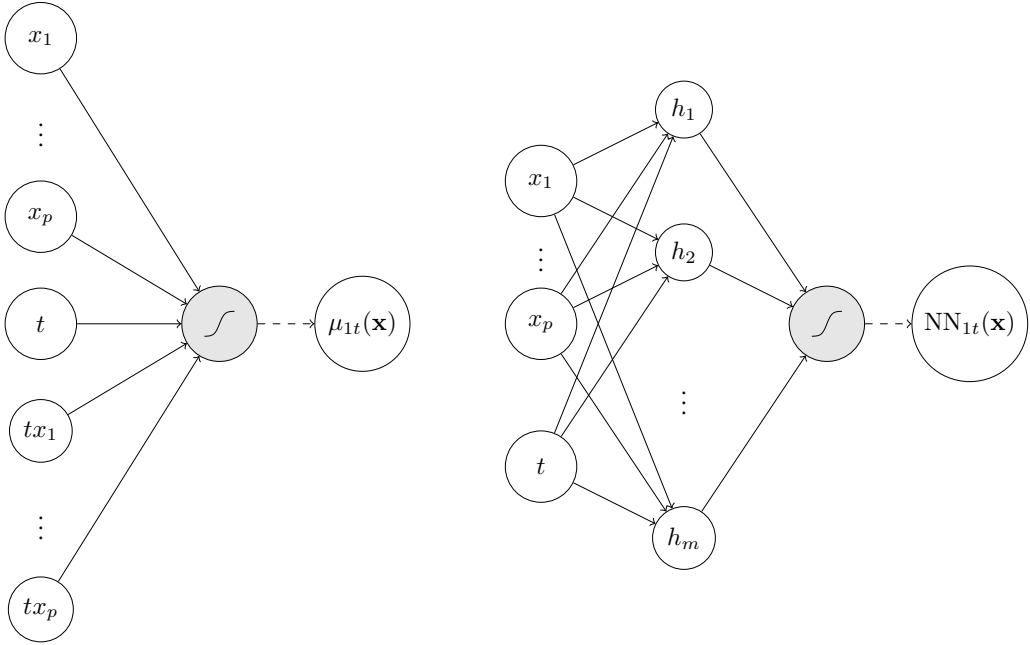
\begin{figure}[!ht]
    \centering
    \scalebox{0.95}{\begin{tikzpicture}

    \tikzset{dist/.style={path picture= {
    \begin{scope}[x=1pt,y=10pt]
      \draw plot[domain=-6:6] (\x,{1/(1 + exp(-\x))-0.5});
    \end{scope}
    }}}
    \tikzstyle{sigmoid}=[draw,fill=gray!20,circle,minimum size=30pt,inner sep=0pt,dist]
    
    \node(x11) at (0,3) [circle, draw, minimum size=1cm] {$x_1$};
    \node at (0,1.75) {$\vdots$};
    \node(x1p) at (0,0.5) [circle, draw, minimum size=1cm] {$x_p$};
    \node(t1) at (0,-1) [circle, draw, minimum size=1cm] {$t$};
    \node(t1x11) at (0,-2.5) [circle, draw] {$t x_1$};
    \node at (0,-3.75) {$\vdots$};
    \node(t1x1p) at (0,-5) [circle, draw] {$t x_p$};
   
    \node (p1)[sigmoid] at (2.5, -1) {};
    
    \draw[->] (x11) -- (p1) node [midway,above] {};
    \draw[->] (x1p) -- (p1) node [midway,above] {};
    \draw[->] (t1) -- (p1) node [midway,above] {};
    \draw[->] (t1x11) -- (p1) node [midway,above] {};
    \draw[->] (t1x1p) -- (p1) node [midway,above] {};
    
    \node(mu1t) at (4.5,-1) [circle, draw] {$\mu_{1t}(\x)$};
    \draw[->, dashed] (p1) -- (mu1t);
    
    \node(x11) at (7,1) [circle, draw, minimum size=1cm] {$x_1$};
    \node at (7,0) {$\vdots$};
    \node(x1p) at (7,-1) [circle, draw, minimum size=1cm] {$x_p$};
    \node(t1) at (7,-3) [circle, draw, minimum size=1cm] {$t$};

    \node(h11) at (9,2) [circle, draw] {$h_1$};
    \node(h12) at (9,0) [circle, draw] {$h_2$};
    \node at (9,-2) {$\vdots$};
    \node(h1m) at (9,-4) [circle, draw] {$h_m$};
    
    \node (p1)[sigmoid] at (11, -1) {};

    \draw[->] (x11) -- (h11) node [midway,above] {};
    \draw[->] (x1p) -- (h11) node [midway,above] {};
    \draw[->] (t1) -- (h11) node [midway,above] {};
    
    \draw[->] (x11) -- (h12) node [midway,above] {};
    \draw[->] (x1p) -- (h12) node [midway,above] {};
    \draw[->] (t1) -- (h12) node [midway,above] {};
    
    \draw[->] (x11) -- (h1m) node [midway,above] {};
    \draw[->] (x1p) -- (h1m) node [midway,above] {};
    \draw[->] (t1) -- (h1m) node [midway,above] {};

    \draw[->] (h11) -- (p1) node [midway,above] {};
    \draw[->] (h12) -- (p1) node [midway,below] {};
    \draw[->] (h1m) -- (p1) node [midway,below] {};

    \node(mu1t) at (13,-1) [circle, draw] {$\mathrm{NN}_{1t}(\x)$};
    \draw[->, dashed] (p1) -- (mu1t);

\end{tikzpicture}}
    \caption{Graphical representation of the uplift interaction model (left panel). Neural network representation of the uplift model (right panel). Note that $h_k = \mathrm{ReLU}(\x,t)$ for $k=1,\ldots,m$. For $m=2p+1$,  Theorem~\ref{theorem} shows the equivalence between the interaction model \eqref{eq:neural_network_1} and a particular case of the $1$-hidden layer neural network \eqref{eq:neural_network_2} with  $\mathrm{ReLU}$ activation. In both diagrams, the gray node represents the sigmoid activation function.}
    \label{fig:int_as_nn}
\end{figure}

Logistic regression may be visualised in a model diagram, with a single node to represent the link function, and multiple nodes to represent inputs or outputs. This sort of visualization is very common in neural networks community (see Figure~\ref{fig:int_as_nn}, left panel). 

The uplift interaction model can be represented by a fully-connected neural network with no hidden layer, an intercept, $2p+1$ input neurons (covariates, treatment variable and interaction terms) and $1$ output neuron with sigmoid activation function, where $\sigma(z)=1/(1+e^{-z})$, for $z \in \real$ (see Figure~\ref{fig:int_as_nn} left panel). Let $\x = (x_1,...,x_p) \in \real^{p}$ be the covariates vector and $t \in \{0,1\}$, a binary variable. Let us further define $\theta_{j}$, for $j=1,\ldots,2p+1$, the coefficient or weight that connects the $j$th input neuron to the output and let $\theta_o \in \mathbb{R}$ be the intercept.
The uplift interaction model can be written as
\begin{equation}
    \mu_{1t}(\x, \boldsymbol{\theta}) = \sigma \Big(\theta_o + \sum_{j=1}^{p} \theta_j x_j + \sum_{j=p+1}^{2p} \theta_j t x_{j-p} +  \theta_{2p+1} t  \Big), ~~ t \in \{0,1\},
    \label{eq:neural_network_1}
\end{equation}
where $\sigma(\cdot)$ represents the sigmoid function and $\boldsymbol{\theta}$ denotes the vector of model parameters. The predicted uplift associated with the covariates vector $\mathbf{x}_{n+1}$ of a future individual is 
$$ \hat{u}(\mathbf{x}_{n+1}) = \mu_{11}(\x_{n+1}, \boldsymbol{\hat{\theta}}) - \mu_{10}(\x_{n+1}, \boldsymbol{\hat{\theta}}),$$
where $\boldsymbol{\hat{\theta}}$ may be estimated by minimizing a loss function such as the one defined in Equation \eqref{eq:augmented_loss_function}.

More generally, let $\mathrm{NN}_{1t}(\x, \boldsymbol{\theta})$ for $t \in \{0,1\}$ be a neural network. We denote by $\mathrm{NN}_{11}(\x, \boldsymbol{\theta})$ and $\mathrm{NN}_{10}(\x, \boldsymbol{\theta})$ the conditional mean model for treated and control observations respectively. In what follows, our goal is to generalize the interaction model \eqref{eq:neural_network_1} by a more flexible neural network. We focus on a fully-connected network with an input of size $p+1$ (covariates and treatment variable), and one hidden layer of size $m>1$ with ReLU activation, where $\mathrm{ReLU}(z) = \max\{0,z\}$, for $z \in \real$. We assume that the intercept (also called bias term) is inherent in the neural model. The hidden layer is then connected to a single output neuron with a sigmoid activation (see Figure~\ref{fig:int_as_nn} right panel). The output of the neural network $\mathrm{NN}_{1t}(\x, \boldsymbol{\theta})$ can be written as
\begin{equation}
    \mathrm{NN}_{1t}(\x, \boldsymbol{\theta}) = \sigma \biggl\{ \theta_o^{(2)} + \sum_{k=1}^{m} \theta_k^{(2)}  \mathrm{ReLU}\Big(\theta_{o,k}^{(1)} + \sum_{j=1}^{p} \theta_{j,k}^{(1)} x_j +  \theta_{p+1,k}^{(1)} t \Big)  \biggr\}, ~~ t \in \{0,1\},
    \label{eq:neural_network_2}
\end{equation}
where $\theta_{j,k}^{(1)}$ represent the coefficient or weight that connects the $j$th covariate or input neuron to the $k$th hidden neuron and $\theta_k^{(2)}$ represents the coefficient that connects the $k$th hidden neuron to the output. We denote the bias terms for the hidden layer and the output layer by $\theta_{o,k}^{(1)}$, $k=1,\ldots,m$  and $\theta_o^{(2)}$ respectively. Here, $\boldsymbol{\theta}$ contains all of the neural network's coefficients (or parameters). The predicted uplift associated with the covariates vector $\mathbf{x}_{n+1}$ of a future individual is 
$$ \hat{u}(\mathbf{x}_{n+1}) = \mathrm{NN}_{11}(\x_{n+1}, \boldsymbol{\hat{\theta}}) - \mathrm{NN}_{10}(\x_{n+1}, \boldsymbol{\hat{\theta}}),$$
where $\boldsymbol{\hat{\theta}}$ may be estimated by minimizing a loss function such as the one defined in Equation \eqref{eq:augmented_loss_function}. In the following Theorem, we show that for a judicious choice of the neural network's coefficients matrix, the two models are equivalent.\\

\begin{theorem} \label{theorem}
Let $\mu_{1t}(\x)$ and $\mathrm{NN}_{1t}(\x)$ be two uplift models defined as in \eqref{eq:neural_network_1} and \eqref{eq:neural_network_2} respectively. Let $c \in \real^+$ be a positive and finite constant and $m=2p+1$. For all $\theta_j \in \real$, $j=1,\ldots,2p+1$ and $\theta_o \in \real$, there exists a matrix of coefficients $\Big(\theta_{j,k}^{(1)}\Big) \in \real^{(p+1) \times (2p+1)}$, such as
\[ \Big(\theta_{j,k}^{(1)}\Big) = 
\left(
\begin{array}{cccc|cccc|c}
1 & 0 & \cdots & 0 & 1 & 0 & \cdots & 0 & 0  \\
0 & 1 & \cdots & 0 & 0 & 1 & \cdots & 0 & 0  \\
\vdots  & \vdots  & \ddots & \vdots & \vdots  & \vdots  & \ddots & \vdots & \vdots  \\
0 & 0 & \cdots & 1 & 0 & 0 & \cdots & 1 & 0  \\
\hline
0 & 0 & \cdots & 0 & c & c & \cdots & c & 1  \\
\end{array}
\right),
\]
an intercepts vector $\Big(\theta_{o,k}^{(1)}\Big) \in \real^{2p+1}$, a vector of coefficients $\Big(\theta_{k}^{(2)}\Big) \in \real^{2p+1}$ and an intercept scalar $\theta_o^{(2)} \in \real$ such that for all $\x \in [0,c]^p$ and $t \in \{0,1\}$
\begin{equation}
    \mu_{1t}(\x) = \mathrm{NN}_{1t}(\x).
    \label{eq:theorem_equality}
\end{equation}
\end{theorem}

\begin{proof}
See Appendix \ref{sec:appendix_thoerem_nn_mu}.
\end{proof}

This theorem is interesting since the neural network model is much more flexible and can be seen as a generalization of the interaction model. As we will see in the experimental Section, this flexibility allows a better fit resulting in a higher performance from a prediction point of view.

\paragraph{A twin representation of uplift models.}

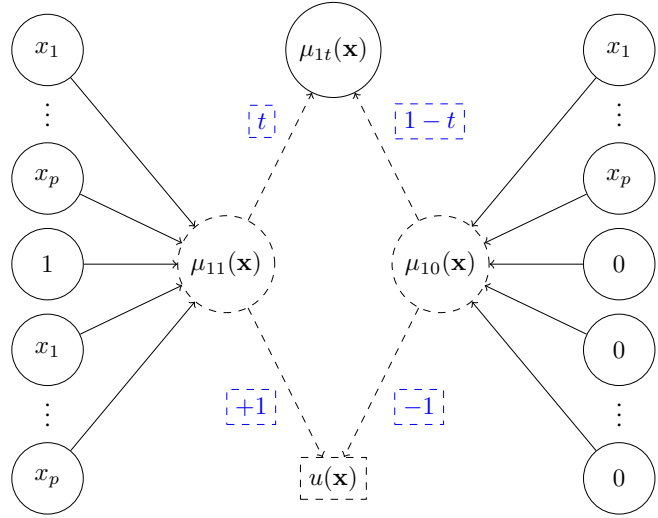
\begin{figure}[!ht]
    \centering
    \scalebox{0.95}{\begin{tikzpicture}

    \tikzset{dist/.style={path picture= {
    \begin{scope}[x=1pt,y=10pt]
      \draw plot[domain=-6:6] (\x,{1/(1 + exp(-\x))-0.5});
    \end{scope}
    }}}
    \tikzstyle{sigmoid}=[draw,fill=gray!20,circle,minimum size=30pt,inner sep=0pt,dist]

    \node(x11) at (0,2) [circle,draw, minimum size=1cm] {$x_1$};
    \node at (0,1.2) {$\vdots$};
    \node(x1p) at (0,0.2) [circle,draw, minimum size=1cm] {$x_p$};
    \node(t1) at (0,-1) [circle,draw, minimum size=1cm] {$1$};
    \node(t1x11) at (0,-2.2) [circle,draw, minimum size=1cm] {$x_1$};
    \node at (0,-3) {$\vdots$};
    \node(t1x1p) at (0,-4) [circle,draw, minimum size=1cm] {$x_p$};
    
    
    \node(x21) at (8,2) [circle,draw, minimum size=1cm] {$x_1$};
    \node at (8,1.2) {$\vdots$};
    \node(x2p) at (8,0.2) [circle,draw, minimum size=1cm] {$x_p$};
    \node(t0) at (8,-1) [circle,draw, minimum size=1cm] {$0$};
    \node(t0x21) at (8,-2.2) [circle,draw, minimum size=1cm] {$0$};
    \node at (8,-3) {$\vdots$};
    \node(t0x2p) at (8,-4) [circle,draw, minimum size=1cm] {$0$};
    
    \node (p1)[] at (2.5, -1) [circle, draw, dashed] {$\mu_{11}(\x)$};
    \node (p0)[] at (5.5, -1) [circle, draw, dashed] {$\mu_{10}(\x)$};
    
    \draw[->] (x11) -- (p1) ;
    \draw[->] (x1p) -- (p1) ;
    \draw[->] (t1) -- (p1) ;
    \draw[->] (t1x11) -- (p1) ;
    \draw[->] (t1x1p) -- (p1) ;

    \draw[->] (x21) -- (p0) ;
    \draw[->] (x2p) -- (p0) ;
    \draw[->] (t0) -- (p0) ;
    \draw[->] (t0x21) -- (p0) ;
    \draw[->] (t0x2p) -- (p0) ;
    
    \node(uplift) at (4,-4) [rectangle, draw, dashed] {$u(\x)$};
    
    \node(mean) at (4,2) [circle, draw] {$\mu_{1t}(\x)$};
    
    \draw[->, dashed] (p1) -- (mean); 
    \node at (3,1) [rectangle, dashed, blue, draw] {$t$};
    \draw[->, dashed] (p0) -- (mean); 
    \node at (5.35,1) [rectangle, dashed, blue, draw] {$1-t$};
    
    \draw[->, dashed] (p1) -- (uplift); 
    \node at (2.85,-3) [rectangle, dashed, blue, draw] {$+1$};
    \draw[->, dashed] (p0) -- (uplift); 
    \node at (5.2,-3) [rectangle, dashed, blue, draw] {$-1$};
    
    \end{tikzpicture}}
    \caption{Diagram of the twin logistic interaction model. The original interaction model is separated into two sub-components with the same parameters. For the left sub-component, the treatment variable fixed to $1$ and the interaction terms to $\x$. The treatment variable and the interactions terms are fixed to $0$ for the right sub-component. The sub-components model the conditional means for treated ($\mu_{11}(\x)$) and for control ($\mu_{10}(\x)$). The difference gives direct prediction of $u(\x)$. At the same time, the predicted conditional mean $\mu_{1t}(\x)$ is based on the actual received treatment for each individual $t \in \{0,1\}$, i.e., $\mu_{1t}(\x) = t \mu_{11}(\x) + (1-t) \mu_{10}(\x)$.}
    \label{fig:siamese_network}
\end{figure}

For most existing parametric uplift methods, the uplift prediction is computed in several steps: i) the uplift model is fitted; ii) the conditional probabilities are predicted by fixing the treatment variable $T$ to $1$ or $0$; iii) the difference is taken to compute the uplift; iv) the uplift is visualized. The fitted model plays a major role in implementing each of these steps. This can be problematic when the fitted model overfit the data at hand. Therefore, most multi-steps methods require careful regularization. To simplify this task, we propose to combine the whole process into a single step through a twin model. The twin interaction model diagram is visualized in  Figure~\ref{fig:siamese_network}. This representation resembles the twin networks \citep{bromley1994signature} in the context of deep learning. Such networks were first introduced in signature verification as an image matching problem, where the task is to compare two hand-written signatures and infer the identity of the writer.

A twin neural network consists of two models that use the same parameters (or weights) while fitted in parallel on two different input vectors to compute comparable outputs. In our case, the input vectors are almost identical, only the treatment variable is changed. The parameters between the twin networks are shared. Weight sharing guarantees that two individuals with similar characteristics are mapped similarly by their respective networks because each network computes the same function. Such networks are mostly known for their application in face recognition \citep{chopra2005learning,taigman2014deepface,parkhi2015deep,schroff2015facenet}, areal-to-ground image matching \citep{lin2015learning} 
and large scale video classification \citep{karpathy2014large}
, among others. 

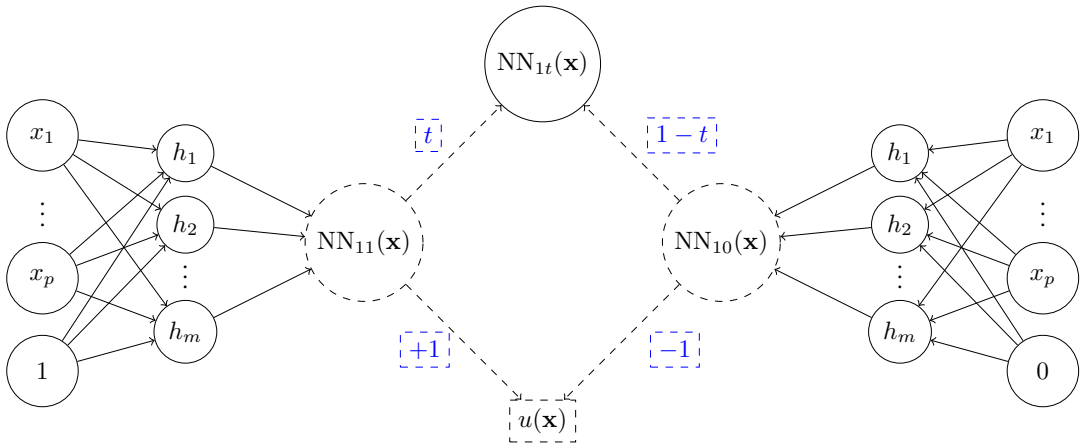
\begin{figure}[H]
    \centering
    \scalebox{0.95}{\begin{tikzpicture}

    \tikzset{dist/.style={path picture= {
    \begin{scope}[x=1pt,y=10pt]
      \draw plot[domain=-6:6] (\x,{1/(1 + exp(-\x))-0.5});
    \end{scope}
    }}}
    \tikzstyle{sigmoid}=[draw,fill=gray!20,circle,minimum size=20pt,inner sep=0pt,dist]

    \node(x11) at (-3,1) [circle, draw, minimum size=1cm] {$x_1$};
    \node at (-3,0) {$\vdots$};
    \node(x1p) at (-3,-1) [circle, draw, minimum size=1cm] {$x_p$};
    \node(t1) at (-3,-2.3) [circle, draw, minimum size=1cm] {$1$};

    \node(x21) at (11,1) [circle, draw, minimum size=1cm] {$x_1$};
    \node at (11,0) {$\vdots$};
    \node(x2p) at (11,-1) [circle, draw, minimum size=1cm] {$x_p$};
    \node(t0) at (11,-2.3) [circle, draw, minimum size=1cm] {$0$};
  
    \node(h11) at (-1,0.75) [circle, draw] {$h_1$};
    \node(h12) at (-1,-0.25) [circle, draw] {$h_2$};
    \node at (-1,-0.875) {$\vdots$};
    \node(h1m) at (-1,-1.75) [circle, draw] {$h_m$};
    \node(h21) at (9,0.75) [circle, draw] {$h_1$};
    \node(h22) at (9,-0.25) [circle, draw] {$h_2$};
    \node at (9,-0.875) {$\vdots$};
    \node(h2m) at (9,-1.75) [circle, draw] {$h_m$};
    
    \node (p1)[] at (1.5, -0.5) [circle, draw, dashed] {$\mathrm{NN}_{11}(\x)$};
    \node (p0)[] at (6.5, -0.5) [circle, draw, dashed] {$\mathrm{NN}_{10}(\x)$};
    
    \draw[->] (x11) -- (h11) node [midway,above] {};
    \draw[->] (x1p) -- (h11) node [midway,above] {};
    \draw[->] (t1) -- (h11) node [midway,above] {};
    
    \draw[->] (x11) -- (h12) node [midway,above] {};
    \draw[->] (x1p) -- (h12) node [midway,above] {};
    \draw[->] (t1) -- (h12) node [midway,above] {};
    
    \draw[->] (x11) -- (h1m) node [midway,above] {};
    \draw[->] (x1p) -- (h1m) node [midway,above] {};
    \draw[->] (t1) -- (h1m) node [midway,above] {};
    
    \draw[->] (x21) -- (h21) node [midway,above] {};
    \draw[->] (x2p) -- (h21) node [midway,above] {};
    \draw[->] (t0) -- (h21) node [midway,above] {};
    
    \draw[->] (x21) -- (h22) node [midway,above] {};
    \draw[->] (x2p) -- (h22) node [midway,above] {};
    \draw[->] (t0) -- (h22) node [midway,above] {};
    
    \draw[->] (x21) -- (h2m) node [midway,above] {};
    \draw[->] (x2p) -- (h2m) node [midway,above] {};
    \draw[->] (t0) -- (h2m) node [midway,above] {};
    
    \draw[->] (h11) -- (p1) node [midway,above] {};
    \draw[->] (h12) -- (p1) node [midway,below] {};
    \draw[->] (h1m) -- (p1) node [midway,below] {};
    
    \draw[->] (h21) -- (p0) node [midway,above] {};
    \draw[->] (h22) -- (p0) node [midway,below] {};
    \draw[->] (h2m) -- (p0) node [midway,below] {};

    \node(uplift) at (4,-3) [rectangle, draw, dashed] {$u(\x)$};
    
    \node(mean) at (4,2) [circle, draw] {$\mathrm{NN}_{1t}(\x)$};
    
    \draw[->, dashed] (p1) -- (mean); 
    \node at (2.4,1) [rectangle, draw, blue, dashed] {$t$};
    \draw[->, dashed] (p0) -- (mean); 
    \node at (5.95,1) [rectangle, draw, blue, dashed] {$1-t$};
    
    \draw[->, dashed] (p1) -- (uplift); 
    \node at (2.35,-2) [rectangle, draw, blue, dashed] {$+1$};
    \draw[->, dashed] (p0) -- (uplift); 
    \node at (5.85,-2) [rectangle, draw, blue, dashed] {$-1$};

    \end{tikzpicture}}
    \caption{A twin neural model for uplift. The inputs contain the covariates vector $\x$ and, for the left sub-component, the treatment variable fixed to $1$. The treatment variable is fixed to $0$ for the right sub-component. The sub-components output the predicted conditional means for treated ($\mathrm{NN}_{11}(\x)$) and for control ($\mathrm{NN}_{10}(\x)$). The difference gives direct prediction of $u(\x)$. At the same time, the predicted conditional mean $\mathrm{NN}_{1t}(\x)$ is based on the actual received treatment for each individual $t \in \{0,1\}$, i.e., $\mathrm{NN}_{1t}(\x) = t \mathrm{NN}_{11}(\x) + (1-t) \mathrm{NN}_{10}(\x)$.}
    \label{fig:nite}
\end{figure}

The twin network  representation of the uplift interaction model is easily generalized to neural networks, as show in Figure~\ref{fig:nite}. In our case, instead of comparing two distinct images, we duplicate each observation and fix the treatment variable to $1$ and $0$, while keeping in memory the true value $t$ of the treatment variable. This makes it possible to fit the twins in parallel and to use the true value $t$ to compute $\mathrm{NN}_{1t}(\x)$. The availability of $\mathrm{NN}_{11}(\x)$ and $\mathrm{NN}_{10}(\x)$ allows to compute the uplift loss function \eqref{eq:augmented_loss_function} and to predict the uplift $u(\x)$ in a single step, which simplifies the fitting process to a great extent.

\section{Parameter estimation}

Loss functions are generally convex with respect to $\mathrm{NN}(\cdot)$, but not with respect to the model's parameters $\boldsymbol\theta$. Parameter estimation is difficult, and with neural networks, in the best cases we are looking for a good local minimum. Most of the methods are based on some form of gradient descent. Gradient-based optimization is one of the pillars of machine learning. Given the  loss function $\ell: \real \rightarrow \real$, classical gradient descent has the goal of finding (local) minima $\boldsymbol{\hat{\theta}} = \argmin\limits_{\boldsymbol{\theta}} \ell(\boldsymbol{\theta})$ via updates of the form $ \Delta (\boldsymbol{\theta}) =  -\eta \nabla \ell$, where $\eta$ is a positive step size (or the learning rate). 

The prototypical stochastic optimization method is the stochastic gradient
method \citep{robbins1951stochastic}, which, in the context of minimizing $\ell(\boldsymbol{\theta})$  with a given starting point  $\boldsymbol{\theta}^{(0)}$  is 

\begin{equation}
    \boldsymbol{\theta}^{(q+1)} \leftarrow \boldsymbol{\theta}^{(q)} - \eta \nabla \ell_{i_q} \left(\boldsymbol{\theta}^{(q)}\right),
\end{equation}
where $\ell_{i_q}(\cdot)$ is the loss function evaluated using observation $i_q\in \{1,...,n\}$ which is chosen at random. Here we use the term stochastic in the sense that for each parameters update, only a random sample of the data is used. This method is to be distinguished from a gradient descent method using a stochastic learning rate $\eta_q$. 

There is no particular reason to employ information from only one observation per iteration. Instead, one can employ a {\it mini-batch} approach in which a small subset of observations $\mathcal{S}_q \subseteq \{1, ..., n\}$, is chosen randomly in each iteration, leading to

\begin{equation}
    \boldsymbol{\theta}^{(q+1)} \leftarrow \boldsymbol{\theta}^{(q)} - \frac{\eta}{|\mathcal{S}_q|} \sum_{i \in \mathcal{S}_q} \nabla \ell_{i} (\boldsymbol{\theta}^{(q)}),
    \label{eq:mini-batch-sgd}
\end{equation}
where $\ell_{i}(\cdot)$ is the loss function evaluated using observation $i$ and $|\mathcal{S}_q|$ is the batch size. Such a mini-batch stochastic gradient method has been widely used in practice.  There are some fundamental practical and theoretical reasons why stochastic methods have inherent advantages for large-scale machine learning \citep{bottou2018optimization}. Note that optimization algorithms that use only the gradient are called first-order optimization algorithms. Second-order optimization algorithms such as Newton's method use the Hessian matrix. Like all neural network training, we estimate the neural network's parameters using first-order gradient methods. 

Neural networks are typically over-parametrized and are prone to overfitting. Hence, regularization is required. Typically the regularization term is a mixture of $L_2$ (Ridge) and $L_1$ (lasso) norms, each with its own regularization constant. Similar to classical regression applications, the intercepts are not penalized. The regularization constants are typically small and serve several roles. The $L_2$ reduces collinearity, and $L_1$ ignores irrelevant parameters, and both are a remedy to the overfitting, especially in over-parametrized models such as deep neural networks \citep{efron2016computer}.

\subsection{Unstructured sparsity}

The parameter estimation process optimizes the uplift loss, so in the interaction model  
\begin{equation}
\boldsymbol{\hat{\theta}} = \argmin_{\boldsymbol{\theta}} \ell(\boldsymbol{\theta}) +\lambda \R(\boldsymbol{\theta}),
\label{eq:optim_uplift_interact}
\end{equation}
where $\ell(\boldsymbol{\theta})$ is the loss, $\lambda \in \real^+$ is the regularization constant and $\R(\boldsymbol{\theta})$ is often convex and probably non-differentiable such as $\norm{\boldsymbol{\theta}}_1$.
The stochastic gradient descent (SGD) for the lasso needs to be modified to make sure the solution remains sparse, so SGD needs to be adapted using the projected gradient update (see \cite{mosci2010solving} for more details). Proximal gradient methods are a generalized form of projection used to solve non-differentiable convex optimization problems. For the reader wishing to have more knowledge on proximal operators, we suggest \cite[Chapter 6]{beck2017first}.

Let's focus only on a univariate $\theta_j$. We adjust SGD by splitting $\theta_j$ into two positive components, $u_j=\theta_j^+$ and $v_j=\theta_j^-$. Here, $\theta_j^+$ and $\theta_j^-$ are the positive and negative parts of $\theta_j \in \real$ so $\theta_j= u_j-v_j$ in which $u_j,v_j\geq 0$ and of course $|\theta_j| = u_j+v_j$. The optimization problem \eqref{eq:optim_uplift_interact}  can be reformulated as 
$$\argmin_{\mathbf u,\mathbf v} \ell(\mathbf u-\mathbf v ) +\lambda \mathbf 1^\top(\mathbf u+\mathbf v).$$
Let's focus only on a single update of the parameter $\theta_j=u_j-v_j$ and a single batch size. 
The optimization routine ensures $u_j\geq 0, v_j\geq 0$, by passing the optimizing parameters $\mathbf u, \mathbf v$ through the proximal projection $\mathrm{ReLU}(\cdot)$  to ensure non-negativity.  

Our proximal gradient descent has only two basic steps which are iterated until convergence. For given  $u_j^{(0)}, v_j^{(0)}$ the modified SGD is
\begin{enumerate}
    \item \emph{gradient step}: define intermediate points $\Tilde{u}_j^{(q)}, \Tilde{v}_j^{(q)}$ by taking a gradient step such as
    \begin{align}
        \Tilde{u}_j^{(q)} &= u_j^{(q)} - \eta \{ \lambda + \nabla \ell_{i_q}(\theta_j^{(q)})\}\nonumber\nonumber\\
        \Tilde{v}_j^{(q)} &= v_j^{(q)} - \eta \{ \lambda - \nabla \ell_{i_q}(\theta_j^{(q)})\} 
                \label{eq:lasso_prox_update1}
    \end{align}
    \item \emph{projection step}: evaluate the proximal operator at the intermediate points $\Tilde{u}_j^{(q)}, \Tilde{v}_j^{(q)}$ such as
    \begin{align}
        u_j^{(q+1)} &\leftarrow \mathrm{ReLU}(\Tilde{u}_j^{(q)})\nonumber\\
        v_j^{(q+1)} &\leftarrow \mathrm{ReLU}(\Tilde{v}_j^{(q)})\nonumber\\
        \theta_j^{(q+1)} &\leftarrow u_j^{(q+1)} - v_j^{(q+1)}  
        \label{eq:lasso_prox_update2}
    \end{align}
\end{enumerate}
Exact zero weight updates appear when  $u_j<0$ and $v_j<0$ to enable unstructured sparsity. 

\subsection{Structured sparsity}

Group lasso is a generalization of the lasso method, when features are grouped into disjoint blocks with a total of $G<p$ groups \citep{yuan2006model}. The formulation of group lasso allows us to define what may constitute a suitable group. For pruning the entire hidden nodes, it is enough to define the block to be the weights that define each node. Take the $1$-hidden layer uplift neural network $\mathrm{NN}_{1t}(\x, \boldsymbol{\theta})$ defined in \eqref{eq:neural_network_2}. The neural model can be rewritten as
\begin{equation}
    \mathrm{NN}_{1t}(\x, \boldsymbol{\theta}) = \sigma \biggl\{ \theta_o^{(2)} + \sum_{k=1}^{m} \theta_k^{(2)}  \mathrm{ReLU}\Big(s_k \{\theta_{o,k}^{(1)} + \sum_{j=1}^{p} \theta_{j,k}^{(1)} x_j +  \theta_{p+1,k}^{(1)} t \} \Big)  \biggr\},
    \label{eq:neural_network_2_scaling_factor}
\end{equation}
where here, we introduced a scaling factor $s_k  \in \real$ for $k=1,\ldots,m$ (i.e., one scaling factor per hidden node). Introduction of a scaling factor for structured pruning is common in neural networks, see for instance \cite{ramakrishnan2020differentiable}.
Let $\mathbf s = (s_1, \ldots, s_m)$ be the $m$-dimensional scaling factors vector. We propose the following optimization problem to enforce structured sparsity in the node level
$$\argmin_{\mathbf u, \mathbf v, \mathbf a, \mathbf b} \ell(\mathbf u, \mathbf v, \mathbf a, \mathbf b) + \lambda_1 \mathbf 1^\top({\mathbf a + \mathbf b}) + \lambda_2 \R(\mathbf u-\mathbf v),$$
where $\lambda_1, \lambda_2 \in \real^+$ are the regularization constants and $
\lambda_1 $ controls the amount of structured sparsity. This formulation allows to use a similar lasso proximal SGD development using the introduced scaling factor $s_k$. So we define $s_k= a_k-b_k$, for $k=1,\ldots,m$ in which $a_k,b_k\geq 0$ so that $|s_k| = a_k+b_k$ and $\lambda_1 \norm{\mathbf s}_1 = \lambda_1 \mathbf 1^\top(\mathbf a+\mathbf b)$. This results in the following modified SGD updates for
given  $a_k^{(0)}, b_k^{(0)}$ in addition to the lasso proximal updates \eqref{eq:lasso_prox_update1} and \eqref{eq:lasso_prox_update2}
\begin{enumerate}
    \item \emph{gradient step}: define intermediate points $\Tilde{a}_k^{(q)}, \Tilde{b}_k^{(q)}$ by taking the gradient step 
    \begin{align*}
        \Tilde{a}_k^{(q)} &= a_k^{(q)} - \eta \{ \lambda_1 + \nabla \ell_{i_q}(s_k^{(q)})\}\\
        \Tilde{b}_k^{(q)} &= b_k^{(q)} - \eta \{ \lambda_1 - \nabla \ell_{i_q}(s_k^{(q)})\}
    \end{align*}
    \item \emph{projection step}: evaluate the proximal operator at the intermediate points $\Tilde{a}_k^{(q)}, \Tilde{b}_k^{(q)}$ such as
    \begin{align*}
        a_k^{(q+1)} &\leftarrow \mathrm{ReLU}(\Tilde{b}_k^{(q)})\\
        b_k^{(q+1)} &\leftarrow \mathrm{ReLU}(\Tilde{a}_k^{(q)})\\
        s_k^{(q+1)} &\leftarrow a_k^{(q+1)} - b_k^{(q+1)} 
    \end{align*}
\end{enumerate}
Pruning the $k$th node happens when $a_k<0$ and $b_k<0$ which enables structured sparsity. This  structured sparsity yields an automatic selection of the number of hidden nodes $m$.


\section{Model Evaluation}\label{sec:qini}

In the context of model selection, in practice, given $L$ models and/or hyperparameter settings, we build $L$ estimators $\{ \hat{u}_1, \hat{u}_2, \ldots, \hat{u}_L \}$. We intend to maximize expected prediction performance, using some goodness-of-fit statistic.  There are several ways to evaluate prediction performance. However, data-splitting to \emph{training data} and \emph{validation data} is the most widely used in practice \citep{arlot2010survey}. Before fitting the models, the observations are randomly split into training samples $\mathcal T$and validation samples $\mathcal V$. All models are fit on  $\mathcal{T}$, but evaluated on $\mathcal{V}$.

Classic evaluation approaches are ineffective  for treatment effect estimation, because both treatment and control are not observed in any observation so the true treatment effect is never observed.
{\it Qini coefficient}, which is based on the {\it Qini curve} \citep{radcliffe2007using}, is commonly used in the uplift literature as an alternative to the goodness-of-fit statistic (see Definition~\ref{eq:q:hat}). The Qini curve separates  observations into heterogeneous segments in terms of  reactions to the treatment and identify sub-groups with most varying predicted uplifts (see Definition~\ref{def:qini_curve}).\\

\begin{definition}
    Given a model, let $\hat{u}_{(1)} \geq \hat{u}_{(2)} \geq ... \geq \hat{u}_{(|\mathcal{V}|)}$ be the sorted predicted uplifts on the validation set $\mathcal{V}$. Let $\phi \in [0,1]$ be a given proportion. Define $ N_{\phi} = \{i: \hat{u}_{i} \geq \hat{u}_{(\ceil{\phi |\mathcal{V}|})} \} \subset \lbrace 1, \ldots, |\mathcal{V}| \rbrace$ as the subset of observations with the $\phi |\mathcal{V}| \times 100 \%$ highest predicted uplifts. 
    The \textit{Qini curve} is the function $g$ of the fraction of population treated $\phi$, where 
\begin{equation}
    g(\phi) = \biggl(\sum\limits_{i \in  N_{\phi}} y_i t_i - \sum\limits_{i \in  N_{\phi}} y_i (1-t_i) \biggl\{ \sum\limits_{i \in  N_{\phi}} t_i / \sum\limits_{i \in  N_{\phi}} (1-t_i) \biggr\} \biggr) / \sum_{i=1}^{|\mathcal{V}|} t_i.
\end{equation}
    \label{def:qini_curve}
\end{definition}

In practice, the domain of $\phi \in [0,1]$ is partitioned into $J$ bins, or $J+1$ grid points $0=\phi_1 < \phi_2 < ... < \phi_{J+1} = 1$. The Qini coefficient is an approximation of the area under the Qini curve.\\

\begin{definition}
    The {\it Qini coefficient} is given by:
\begin{equation}
  \hat{q} = \dfrac{1}{2} \sum_{k=1}^J (\phi_{k+1}-\phi_k)\{Q(\phi_{k+1}) + Q(\phi_{k})\} \times 100\%.
  \label{eq:q:hat}
\end{equation}
where $Q(\phi) = g(\phi) - \phi~g(1)$ and $g(1)$ is the average treatment effect in the validation set.
\end{definition}


Unlike the area under the ROC curve, $\hat{q}$ may take negative values. A negative $\hat q$ means the uplift is worse than random targeting. A good uplift model groups the individuals in decreasing uplift bins. This can be measured by the similarity between the theoretical uplift percentiles of  predictions compared with empirical percentiles observed in the data. Maximizing the \textit{adjusted Qini coefficient}, given in Definition~\ref{def:qadj}, maximizes the Qini coefficient and simultaneously promotes grouping the individuals in decreasing uplift bins, which in turn result in concave Qini curves \citep{belba2019qbased}.\\

\begin{definition}\label{def:qadj}
Let $B_k$ denote the  $k$th bin $(\phi_{k}, \phi_{k+1}] \subseteq (0,1]$, $k=1,\ldots, J$. The \textit{adjusted Qini coefficient} is defined as:
    \begin{equation}
        \qadj = \rho~\mathrm{max}\{ 0, \hat{q}\},
    \label{eq:qadj}
    \end{equation}
where $\rho$ is the {\it Kendall's uplift rank correlation}:
    \begin{equation}
        \hat{\rho} = \frac{2}{K(K-1)} \sum_{i<j} \mathrm{sign}(\bar{\hat{u}}_i - \bar{\hat{u}}_j)~\mathrm{sign}(\bar{u}_i - \bar{u}_j),
    \label{eq:corr_coeff}
    \end{equation}
where $\bar{\hat{u}}_k$ is the average predicted uplift in bin $B_k$,  $k \in {1,...,J}$, and $\bar{u}_k$ is the observed uplift in the same bin.
\end{definition}
For the remainder of the paper, we use $\qadj$ as the models comparison measure.

\section{Experiments}

We demonstrate the utility of our proposed methods in a simulation study. Each simulation is defined by a data-generating process with a known effect function. Each run of each simulation generates a dataset, and split into training, validation, and test subsets. We use the training data to estimate $L$ different uplift functions $\{\hat{u}_l\}_{l=1}^L$ using $L$ different methods. The models are fine-tuned using the training and validation observations and results are presented for the test set.

\subsection{Data generating process}

We generate synthetic data similar  to  \cite{powers2018some}. For each experiment, we generate $n$ observations and $p$ covariates. We draw odd-numbered covariates independently from a standard Gaussian distribution. Then, we draw even-numbered covariates independently from a Bernoulli distribution with probability $1/2$. Across all experiments, we define the mean effect function $\mu(\cdot)$ and the treatment effect function $\tau(\cdot)$ for a given noise level $\sigma^2$. Given the elements above, our data generation model is, for $i=1,...,n$,
\begin{align*}
    &Y_i^* \mid \x_i, t_i \sim \mathcal{N}(\mu(\x_i) + t_i\tau(\x_i), \sigma^2),\\
    &Y_i = \mathbbm{1}(Y_i^* > 0 \mid \x_i, t_i)
\end{align*}
where $t_i$ is the realisation of the random variable $T_i \sim \mathrm{Bernoulli}(1/2)$ and $Y_i$ is the binary outcome random variable. Following \cite{powers2018some},  within each set of simulations, we make different choices of mean effect function and treatment effect function. In this section, we describe the results associated with the most complex scenario since the conclusions are similar from one scenario to another. For the reader wishing to analyze the results for other scenarios, we present them in Appendix~\ref{sec:appendix_experiments}. We fix $n=20000$, $p=100$ and $\sigma=4$ and define $\mu(\x)$ and $\tau(\x)$ as follows
\begin{align*}
    \mu(\x) &= 4 \mathbbm{1}(x_1 > 1)\mathbbm{1}(x_3 > 0) + 4\mathbbm{1}(x_5 > 1)\mathbbm{1}(x_7 > 0) + 2x_8x_9,\\
   \tau(\x) &= \frac{1}{\sqrt{2}} (f_4(\x) + f_5(\x)),\\
\end{align*}
where
\begin{align*}
    f_4(\x) &= x_2 x_4 x_6 + 2 x_2 x_4 (1-x_6) +3x_2(1-x_4)x_6 + 4x_2(1-x_4)(1-x_6) + 5(1-x_2)x_4x_6 \\
            &+ 6(1-x_2)x_4(1-x_6) + 7 (1-x_2)(1-x_4)x_6 + 8(1-x_2)(1-x_4)(1-x_6),\\
    f_5(\x) &= x_1 + x_3 + x_5 + x_7 + x_8 + x_9 - 2.\\
\end{align*}

Next, we repeat each experiment $20$ times and each run generates a dataset, which is divided into training ($40\%$), validation ($30\%$), and test samples ($30\%$). The models are fitted using the training observations and results are presented for the test set.

\subsection{Regularization} 

In Table~\ref{tab:l1_l2_gl_}, we compare the performance of the twin model $\mathrm{NN}_{1t}(\x, \boldsymbol{\theta})$ (defined in \eqref{eq:neural_network_2_scaling_factor}) using different regularization functions, with and without the scaling factors for pruning (i.e., structured sparsity). Next, we will refer to this model as {\it Twin$_{\mathrm{NN}}$}. We fix the initial number of hidden neurons (or nodes) to $m=512 $. When structured sparsity regularization is used, we denote by $\hat m$ the number of remaining hidden neurons. We run the experiments $20$ times.  For fitting the models in each experiment, we vary the hyper-parameters from the following grid: \textit{learning rate} $\eta \in \{ 0.005, 0.01, 0.05, 0.1, 0.2, 0.3 \}$, \textit{structured sparsity constant} $\lambda_1 \in \{0, 0.0001, 0.0005, 0.001, 0.005, 0.01 \}$, \textit{regularization constant} $\lambda_2 \in \{0, 0.0001, 0.0005, 0.001, 0.005, 0.01 \}$ and cross-validate on the $\qadj$ for each combination.

\begin{table}[!ht]
    \centering
    \begin{tabular}{llcc}
        \hline
        Structured Sparsity & $\R(\cdot)$ & $\hat{m}/m$ & $\qadj$ \\
        \hline
        No & $L_2$ & $512/512$ & $3.15$ \\
        No &$L_1$ &  $512/512$ & $3.35$ \\
        Yes & 0 & $218/512$ & $3.09$ \\
        Yes &$L_2$  & $417/512$ & $3.32$ \\
        Yes &$L_1$  & $340/512$ & $\bf 3.58$ \\
        \hline
    \end{tabular}
    \vspace{2mm}
    \caption{Average adjusted Qini (20 runs) for the twin model \eqref{eq:neural_network_2_scaling_factor} with structured pruning of nodes and different regularization functions $\R(\cdot)$. The $L_1$ regularization of weights provides the highest performance. Note that the maximum standard-error is $0.1$; we do not report them to simplify the Table. 
    }
    \label{tab:l1_l2_gl_}
\end{table}

With the $ L_1 $ regularization, the test-set adjusted Qini coefficient is higher than when the $ L_2 $ regularization is used. When structured sparsity is used alone, the model removes too many neurons (around half) but $\qadj$ decreases to $3.09$. On the other hand, when structured sparsity is used in addition to the $ L_2 $ regularization, we see a clear improvement over $L_2$ or structured sparsity alone, and in this case, the number of pruned neurons is smaller. However, it does not outperform the $L_1$ regularization used alone. Finally, the best combination is the $ L_1 $ regularization paired with the structured sparsity achieving an average adjusted Qini coefficient of $3.58$. 

Now, when structured sparsity is used to prune the hidden nodes, it is possible to refit a new model with $\hat{m}$ hidden nodes. In our experiments, this did not necessarily have a positive impact on prediction performance. In the case where only structured sparsity is used, $\hat{m} = 218$. For a model with $ 218 $ hidden neurons fitted with the $ L_2 $ regularization, $\qadj$ drops to $ 2.80 $. With $ L_1 $, it increases slightly to reach an average of $ 3.16 $. The same results are observed when we first use $ L_2 $ and structured sparsity and then refit models with $417$ hidden neurons. Finally, in the case where we fit the models directly with $ L_1 $ and structured sparsity, the prediction performance is at its maximum. Indeed, when we refit the models with $ \hat{m} = 340 $ hidden nodes, whether with $ L_2 $ or $ L_1 $, the adjusted Qini decreases from $3.58$ to $ 3.01 $ and $ 3.31 $ respectively. Other scenarios not presented here yielded similar conclusions. Therefore, it seems better to fit the models in one step, using the $ L_1 $ regularization and structured sparsity. This make it possible to prune a few nodes from the hidden layer, and to get sparse estimation of the remaining parameters. This seems to improve the prediction performance of the underlying model. For the rest of the paper, we fit our models with $ L_1 $ and structured sparsity. 

\subsection{Comparison with benchmark models}
In this section, we compare our models to different benchmarks. First, let us focus on the simple twin model $\mu_{1t}(\x, \boldsymbol{\theta})$, defined in \eqref{eq:neural_network_1}. The goal is to compare three optimization procedures. The first method optimizes the penalized Binomial likelihood, as in the case of a {\it lasso} logistic regression. This is the baseline model, and we will refer to it as \textit{Logistic} as a reference to Lo's interaction model \citep{lo2002true}. For fair comparison, we use the twin neural architecture. The second method is a two-stage method which 
is based on a derivative-free optimization of the $\qadj$ and imposes sparsity \citep{belba2019qbased}. 
We denote this model by \textit{Qini-based} and our proposed method by {\it Twin$_{\mu}$}. We also compare to lasso versions of the R-learner \citep{NieWager2020Quasi} and X-learner \citep{kunzel2019metalearners}. Table~\ref{tab:param_estimation_results_} shows the averaged results on the test dataset. This experience shows the superiority of using our uplift loss function. Results for other scenarios are presented in Appendix~\ref{sec:appendix_experiments}.\\

\begin{table}[!ht]
    \centering
    \begin{tabular}{ccccc}
        \hline
        {\it Logistic} & {\it Qini-based} & {\it Twin$_{\mu}$} & {\it R-Learner }(\texttt{lasso}) & {\it X-Learner }(\texttt{lasso}) \\
        \hline
         $2.59$ & $2.94$ & $\bf{3.12}$ & $2.83$ & $2.91$\\
         \hline
    \end{tabular}
    \vspace{2mm}
    \caption{``Simple" models prediction performance comparison in terms of $\qadj$. Note that the maximum standard-error is $0.05$; we do not report them to simplify the Table.}
    \label{tab:param_estimation_results_}
\end{table}


Next we compare our twin model $\mathrm{NN}_{1t}(\x, \boldsymbol{\theta})$, defined in \eqref{eq:neural_network_2_scaling_factor}, with commonly used benchmark models. A method that has proven to be very effective in estimating treatment effects is one based on generalized random forests \citep{athey2019generalized}. This method uses honest estimation, i.e., it does not use the same information for the partition of the covariates space and for the estimation of the uplift. This has the effect of reducing overfitting. Another candidate that we consider in our experiments is also based on random forests and designed for uplift \citep{guelman2012random}. For this method, we consider two different split criteria, one based on the Kullback-Leibler (KL) divergence and one based on the Euclidean distance (ED). We also compare our method to XGboost \citep{chen2015xgboost} versions of the R- and X-learners. In Table~\ref{tab:beyond_linearity_results_}, we compare these benchmark models to different versions of our {\it Twin$_{\mathrm{NN}}$} with different values of $m$ (hidden neurons), with and without structured sparsity. Appendix~\ref{sec:appendix_experiments} provides details about model fine-tuning and other scenarios results. \\


\begin{table}[!ht]
    \centering
    \begin{tabular}{lccc}
        \hline
        Method & Structured Sparsity & Hidden-Layer Size ($\hat m/m$) & $\qadj$ \\
        \hline
        {\it Twin$_{\mathrm{NN}}$} & No & $64/64$   &  $3.10$\\
        {\it Twin$_{\mathrm{NN}}$} & No & $128/128$ &  $3.16$\\
        {\it Twin$_{\mathrm{NN}}$} & No & $201/201$ &  $3.14$\\
        {\it Twin$_{\mathrm{NN}}$} & No & $256/256$ &  $3.26$\\
        {\it Twin$_{\mathrm{NN}}$} & No & $512/512$ &  $3.35$\\
        {\it Twin$_{\mathrm{NN}}$} & Yes & $340/512$ & $\bf 3.58$\\
        \hline
        \multicolumn{3}{l}{{\it Causal Forest}}& $2.79$ \\
        \multicolumn{3}{l}{{\it Causal Forest (Honest)}}& $\bf 3.07$\\
        \multicolumn{3}{l}{{\it Uplift Random Forest (KL)}}&$2.19$\\
        \multicolumn{3}{l}{{\it Uplift Random Forest (ED)}}& $2.33$ \\
        \multicolumn{3}{l}{{\it R-Learner }(\texttt{XGboost})}& $2.12$\\
        \multicolumn{3}{l}{{\it X-Learner }(\texttt{XGboost})}& $2.37$ \\
       \hline
    \end{tabular}
    \vspace{2mm}
    \caption{Models prediction performance comparison in terms of $\qadj$. Note that the maximum standard-error is $0.1$; we do not report them to simplify the Table.}
    \label{tab:beyond_linearity_results_}
\end{table}

Neural networks have two main hyper-parameters that control the architecture or topology of the network: the number of hidden layers and the number of nodes in each hidden layer. The number of nodes can be seen as a hyper-parameter. If we don't use our pruning technique for structured sparsity, we can search for the right number of nodes over a grid. For comparison purposes, we varied the number of nodes using either $m=64, 128, 256$ or $m=512$ nodes. As we can see in Table~\ref{tab:beyond_linearity_results_}, the best model is reached for $m = 512$ (with $\qadj=3.35$). Choosing the number of hidden neurons by cross-validation is very common in practice. This is part of architecture search. However, choosing among a few values does not necessarily mean that all the selected model's neurons are useful. As discussed earlier, the use of the scaling factor penalization on the weight matrix allows to automatically fine-tune the number of hidden neurons. This has the effect of increasing the performance of the underlying models from a predictive point of view, as shown in Table~\ref{tab:l1_l2_gl_}. Thus, we suggest to start with a fairly large number of hidden neurons (e.g., $m \geq 2p +1$), and to let the optimization determine the number of active neurons needed for a specific dataset. 

Based on these experiments, it appears that the twin neural model fitted with $L_1$ regularization and structured sparsity outperforms all other methods significantly in terms of $\qadj$ (e.g., based on  Wilcoxon signed-rank test \citep{wilcoxon1992individual}). We also observe that the random forest with honesty criteria performs best in comparison to the other uplift benchmark models. 

Finally, the number of hidden layers can also be seen as a hyper-parameter for the neural network architecture. Note that in our experiments, increasing the number of hidden layers did not improve the prediction performance significantly. For instance, we fixed the initial number of hidden neurons to $p+1$ and $p$ for a $2$-hidden layers twin model and observed an average $\qadj$ of $3.39$. Results for other scenarios are given in Appendix~\ref{sec:appendix_experiments}.

\section{Application}

We have the rare opportunity to have access to real data from a large scale randomized experiment. Indeed, in the uplift domain, it is not easy to find public benchmarks, apart from the recent CRITEO-UPLIFT1 dataset \citep{Diemert2018}.

The CRITEO-UPLIFT1 dataset is constructed by collecting data from a particular randomized trial procedure where a random part of the population is prevented from being targeted by advertising. The dataset consists of $\approx 14M$ rows, each one representing a user with $p=12$ covariates, a treatment indicator and a binary response variables (visits). Positive responses mean the user visited the advertiser website during the test period (2 weeks). The proportion of treated individuals is $85\%$. We use an undersampling method in order to bring the treatment/control ratio back to 1. Thus, we have about $4M$ observations. We also create two other balanced datasets by randomly sampling $400K$ and $1M$ observations. 

We fit the twin-network {\it Twin$_{\mathrm{NN}}$} and optimize the regularized loss function with lasso and structured sparsity. For comparison purposes, we also consider the honest causal forest, which gave the best benchmark in our simulation study. We split the data into training ($40\%$), validation ($30\%$), and test samples ($30\%$). We use the training and the validation sets to fit and fine-tune the models. Then, we select the best model based on the validation set. Using the best model, we score the samples from the test set and compute the adjusted Qini coefficient $\qadj$. We repeat the experiment $20$ times and report averaged results as well as their standard-errors in Table~\ref{tab:final_models_results2}.\\

\begin{table}[!ht]
    \centering
    \begin{tabular}{llc|cc|c}
        \hline
        Dataset & Sample Size & $p$ &{\it Twin$_{\mathrm{NN}}$} & $\hat{m}/m$ &  {\it Causal Forest (Honest)} \\
        \hline
        Insurance data & 50K & $40$ & $0.19~(0.003)$ &  $212/512$ & $0.06~(0.007)$ \\
        CRITEO-UPLIFT1 & 100K& $12$ & $0.19~(0.008)$ & $187/512$  & $0.14~(0.011)$ \\
        CRITEO-UPLIFT1 & 1M  & $12$ & $0.24~(0.007)$ & $127/512$  & $0.18~(0.019)$ \\
        CRITEO-UPLIFT1 & 4M  & $12$ & $0.28~(0.006)$ & $63/512$   & $0.23~(0.015)$\\
        \hline
    \end{tabular}
    \vspace{2mm}
    \caption{Application on real-data. Prediction performance in terms of $\qadj$ averaged over $20$ realizations (standard-errors are given in parenthesis).}
    \label{tab:final_models_results2}
\end{table}

The same type of analysis is repeated on a data set that was made available to us by a car insurance company with $50K$ customers and $p=40$ covariates. This company was interested in designing strategies to maximize its conversion rate. An experimental acquisition campaign was implemented for $6$ months, for which half of the potential clients were randomly allocated into a treatment group and the other half into a control group. Potential clients under the treatment group were contacted. The goal of the analysis is to propose a predictive model that maximizes the return on investment of future initiatives (i.e., a maximum $\qadj$ on a test set). The observed difference in sales rates between the treated and the control groups shows a slightly positive impact of the marketing initiative, that is, $0.55\%$. Results are reported in Table~\ref{tab:final_models_results2}.


For the insurer's data, we also considered a model without hidden layers (i.e., {\it Twin$_{\mu}$}). The objective being model interpretability. In this case, the model is not as efficient as the neural network one (see Table~\ref{tab:final_models_results}). However, it performs better than the causal forest. Therefore, if the practitioner needs an interpretable model, he can use the simple twin model and still get satisfactory results. For comparison purposes, we also fit the {\it Logistic} model. In this case, the model's parameters are estimated by maximizing the penalized Binomial log-likelihood. 

\begin{table}[!ht]
    \centering
    
    \begin{tabular}{lcccc}
        \hline
        Method & $\hat{m}/m$ &Training & Validation & Test \\
        \hline
        {\it Twin$_{\mu}$}  & - & $0.112~(0.005)$ & $0.098~(0.004)$ & $0.085~(0.005)$\\
        {\it Twin$_{\mathrm{NN}}$} & $212/512$ & $0.223~(0.006)$ & $0.202~(0.003)$ &  $0.187~(0.003)$\\
        \hline
        {\it Logistic} & - & $0.197~(0.029)$  & $0.062~(0.019)$ & $0.021~(0.007)$ \\
        {\it Causal Forest (Honest)} & - & $0.573~(0.028)$  & $0.157~(0.022)$ & $0.059~(0.007)$ \\
        \hline
    \end{tabular}
    \vspace{2mm}
    \caption{Adjusted Qini coefficients on the insurance data (standard errors are given in parenthesis). The results are averaged over $20$ runs.}
    \label{tab:final_models_results}
\end{table}

For these types of marketing initiatives, it makes sense that the overall impact is small. A potential reason is that customers are already interested in the product since it is mandatory to get car insurance. The effect of the call is slightly positive on average, that is to say that during the call, the advisor allows the customer to understand all the details of the coverage and thus the customer is most likely reassured. Therefore, it is rather unlikely that the call would have a negative effect. We believe that for this data set, the situation is similar to the first scenario from the simulations presented in Appendix~\ref{sec:appendix_experiments}. This may explain why it is more difficult to prevent overfitting with the causal forest model.

\section{Conclusion}

We present a meaningful and intuitive twin neural networks architecture for the problem of uplift modeling. We proposed to  estimate the model's parameters by optimizing a new loss function. This loss function is built by leveraging a connection with the Bayesian interpretation of the relative risk. The twin neural network performs well on predictive tasks and overfitting is minimized by using a proper regularization term. We applied our method to synthetic and real-world data and we compared it with the state-of-the-art methods for uplift. We modify the learning algorithm to allow for structured sparse solutions, which significantly helped training uplift models. Our results show that the twin models significantly outperform the common approaches to uplift such as random forests in all scenarios.

Our methodological development has been driven by real data from a large scale random experiment where the treatment is the marketing initiative. The methodology is also readily applicable to other types of datasets, such as randomized clinical trial data, for predicting variability in medical treatment response. In addition to the quality of prediction obtained in optimal settings, we observed that even the one-layer version of the proposed architecture performs well compared to random forest models. In this case, the model can be interpreted in the same way as a logistic regression.
This interpretation is of great importance in practice, since the deployment of an interpretable model minimizes risks of applying an inadequate model and facilitates convincing business owners. This is also true in the medical world, where understanding the variability in response to treatment predictions is of great importance for advancing precision medicine, but not at the cost of model explanation. 



\appendix

\section{Proof of Theorem~\ref{theorem}}
\label{sec:appendix_thoerem_nn_mu}

\begin{proof}
We prove the theorem by finding explicitly the coefficients of $\mathrm{NN}_{1t}(\x)$ that verify the equality \eqref{eq:theorem_equality}. First, we fix the vector of coefficients that connects the hidden layer to the output such as 
$$\theta_{k}^{(2)} = \theta_j,$$ 
for $k=j$, for $k,j=1,\ldots,2p+1$. Moreover, we fix the intercept $\theta_o^{(2)}$ to be equal to the intercept from the first model, that is, $\theta_o^{(2)} = \theta_o$. Therefore, $\mathrm{NN}_{1t}(\x)$ can be written as
\begin{equation}
    \mathrm{NN}_{1t}(\x) = \sigma \biggl\{ \theta_o + \sum_{k=1}^{2p+1} \theta_k  \mathrm{ReLU}\Big(\theta_{o,k}^{(1)} + \sum_{j=1}^{p} \theta_{j,k}^{(1)} x_j +  \theta_{p+1,k}^{(1)} t \Big)  \biggr\}.
    \label{eq:neural_network2_proof}
\end{equation}

Next, we need to define the right structure for the matrix of coefficients $\Big(\theta_{j,k}^{(1)}\Big) \in \real^{(p+1) \times (2p+1)}$ to be able to recover the $\mu_{1t}(\x)$ model. First, let us fix the intercepts vector $\Big(\theta_{o,k}^{(1)}\Big) \in \real^{2p+1}$ such as
\[ \theta_{o,k}^{(1)} = 
    \left\{ \begin{array}{cl}
      0 & \mathrm{if}~k \in \{1,\ldots,p\}\\
      -c & \mathrm{if}~k \in \{p+1, \ldots, 2p\}\\
      0 & \mathrm{if}~k=2p+1 
            \end{array} \right.
\]

Finally, we give the following representation to the matrix of coefficients $\Big(\theta_{j,k}^{(1)}\Big) \in \real^{(p+1) \times (2p+1)}$, that is,
\[ \Big(\theta_{j,k}^{(1)}\Big) = 
\left(
\begin{array}{cccc|cccc|c}
1 & 0 & \cdots & 0 & 1 & 0 & \cdots & 0 & 0  \\
0 & 1 & \cdots & 0 & 0 & 1 & \cdots & 0 & 0  \\
\vdots  & \vdots  & \ddots & \vdots & \vdots  & \vdots  & \ddots & \vdots & \vdots  \\
0 & 0 & \cdots & 1 & 0 & 0 & \cdots & 1 & 0  \\
\hline
0 & 0 & \cdots & 0 & c & c & \cdots & c & 1  \\
\end{array}
\right).
\]

Therefore, based on these attributions, we have the following equalities. For $k \in \{1, \ldots, p\}$, 
\begin{equation}
    \mathrm{ReLU}\Big(\theta_{o,k}^{(1)} + \sum_{j=1}^{p} \theta_{j,k}^{(1)} x_j +  \theta_{p+1,k}^{(1)} t \Big) = \mathrm{ReLU}(x_k) = x_k.
    \label{eq:1st_equality}
\end{equation}

Then, for $k \in \{p+1, \ldots, 2p\}$,
\begin{equation}
    \mathrm{ReLU}\Big(\theta_{o,k}^{(1)} + \sum_{j=1}^{p} \theta_{j,k}^{(1)} x_j +  \theta_{p+1,k}^{(1)} t \Big) = \mathrm{ReLU}(-c + x_{k-p} + ct) = t x_{k-p}.
    \label{eq:2nd_equality}
\end{equation}

In \eqref{eq:2nd_equality}, the outcome depends on the value of $t$ and the second equality comes from the fact that we have
\[ \mathrm{ReLU}(-c + x_{k-p} + ct) = 
    \left\{ \begin{array}{ll}
     \mathrm{ReLU}(x_{k-p} - c) = 0 & \mathrm{if}~t = 0,\\
      \mathrm{ReLU}(x_{k-p}) = x_{k-p} & \mathrm{if}~t=1,
            \end{array} \right.
\]
therefore, $\mathrm{ReLU}(-c + x_{k-p} + ct) = t x_{k-p}$. Finally, for $k=2p+1$, we have
\begin{equation}
    \mathrm{ReLU}\Big(\theta_{o,k}^{(1)} + \sum_{j=1}^{p} \theta_{j,k}^{(1)} x_j +  \theta_{p+1,k}^{(1)} t \Big) = \mathrm{ReLU}(t) = t.
    \label{eq:3rd_equality}
\end{equation}

Hence, the desired result is obtained by simply replacing \eqref{eq:1st_equality}, \eqref{eq:2nd_equality} and \eqref{eq:3rd_equality} in \eqref{eq:neural_network2_proof}.
\end{proof}

\section{Additional experiments}
\label{sec:appendix_experiments}

In this section, we present in more detail the experimental protocol as well as the results associated with different scenarios. Each simulation is defined by a data-generating process with a known effect function. Each run of each simulation generates a dataset, and split into training, validation, and test subsets. We use the training data to estimate $L$ different uplift functions $\{\hat{u}_l\}_{l=1}^L$ using $L$ different methods. The models are fine-tuned using the training and validation observations and results are presented for the test set.\\


We generate synthetic data similar  to  \cite{powers2018some}. For each experiment, we generate $n$ observations and $p$ covariates. We draw odd-numbered covariates independently from a standard Gaussian distribution. Then, we draw even-numbered covariates independently from a Bernoulli distribution with probability $1/2$. Across all experiments, we define the mean effect function $\mu(\cdot)$ and the treatment effect function $\tau(\cdot)$ for a given noise level $\sigma^2$. Given the elements above, our data generation model is, for $i=1,...,n$,
\begin{align*}
    &Y_i^* \mid \x_i, t_i \sim \mathcal{N}(\mu(\x_i) + t_i\tau(\x_i), \sigma^2),\\
    &Y_i = \mathbbm{1}(Y_i^* > 0 \mid \x_i, t_i)
\end{align*}
where $t_i$ is the realisation of the random variable $T_i \sim \mathrm{Bernoulli}(1/2)$ and $Y_i$ is the binary outcome random variable. 

The random variable $Y_i$ follows a Bernoulli distribution with parameter $\mathrm{Pr}(Y_i^* > 0 \mid \x_i, t_i)$ and it is easy to recover the ``true'' uplift $u_i^*$, that is,
\begin{align}
    u_i^* &= \mathrm{Pr}(Y_i^* > 0 \mid \x_i, T_i = 1) - \mathrm{Pr}(Y_i^* > 0 \mid \x_i, T_i = 0) \nonumber \\
    u_i^* &= \Phi \Bigl( \frac{\mu(\x_i) + \tau(\x_i)}{\sigma} \Bigr) - \Phi \Big( \frac{\mu(\x_i)}{\sigma} \Big) \label{eq:sim_true_uplift}
\end{align}
for $i=1,\ldots,n$, where $\Phi(.)$ is the standard Gaussian cumulative distribution function.\\

Following \cite{powers2018some},  within each set of simulations, we make different choices of mean effect function and treatment effect function, each represents a wide variety of functional forms, univariate and multivariate, additive and interactive, linear and  piecewise constant. Table~\ref{tab:scenarios} gives the mean and treatment effect functions for the different randomized simulations.
\begin{table}[!ht]
    \centering
    \begin{tabular}{l|cccc}
    \hline
        & \multicolumn{4}{c}{Scenarios} \\
        \hline
        Parameters & 1 & 2 & 3 & 4 \\
        \hline
        $n$ & 10000 & 20000 & 20000 & 20000 \\
        $p$ & 200 & 100 & 100 & 100 \\
        $\mu(\x)$ & $f_7(\x)$ & $f_3(\x)$ & $f_2(\x)$ & $f_6(\x)$\\
        $\tau(\x)$ & $f_4(\x)$ & $f_5(\x)$ &$ f_7(\x)$ & $f_8(\x)$\\
        $\sigma$ & $1/2$ & $1$ & $1$ & $4$\\
        \hline
    \end{tabular}
    \vspace{2mm}
    \caption{Specifications for the simulation scenarios. The rows of the table correspond, respectively, to the sample size, dimensionality, mean effect function, treatment effect function and noise level.}
    \label{tab:scenarios}
\end{table}

The functions are
\begin{align*}
    f_2(\x) &= 5 \mathbbm{1}(x_1 > 1) - 5\\
    f_3(\x) &= 2 x_1 - 4 \\
    f_4(\x) &= x_2 x_4 x_6 + 2 x_2 x_4 (1-x_6) +3x_2(1-x_4)x_6 + 4x_2(1-x_4)(1-x_6) + 5(1-x_2)x_4x_6 \\
            &+ 6(1-x_2)x_4(1-x_6) + 7 (1-x_2)(1-x_4)x_6 + 8(1-x_2)(1-x_4)(1-x_6)\\
    f_5(\x) &= x_1 + x_3 + x_5 + x_7 + x_8 + x_9 - 2\\
    f_6(\x) &= 4 \mathbbm{1}(x_1 > 1)\mathbbm{1}(x_3 > 0) + 4\mathbbm{1}(x_5 > 1)\mathbbm{1}(x_7 > 0) + 2x_8x_9\\
    f_7(\x) &= \frac{1}{2} (x_1^2 + x_2 + x_3^2 + x_4 + x_5^2 + x_6 + x_7^2 + x_8 + x_9^2 - 11)\\
    f_8(\x) &= \frac{1}{\sqrt{2}} (f_4(\x) + f_5(\x)).
\end{align*}

\begin{figure}[!ht]
    \centering
    \includegraphics[width=0.49\textwidth]{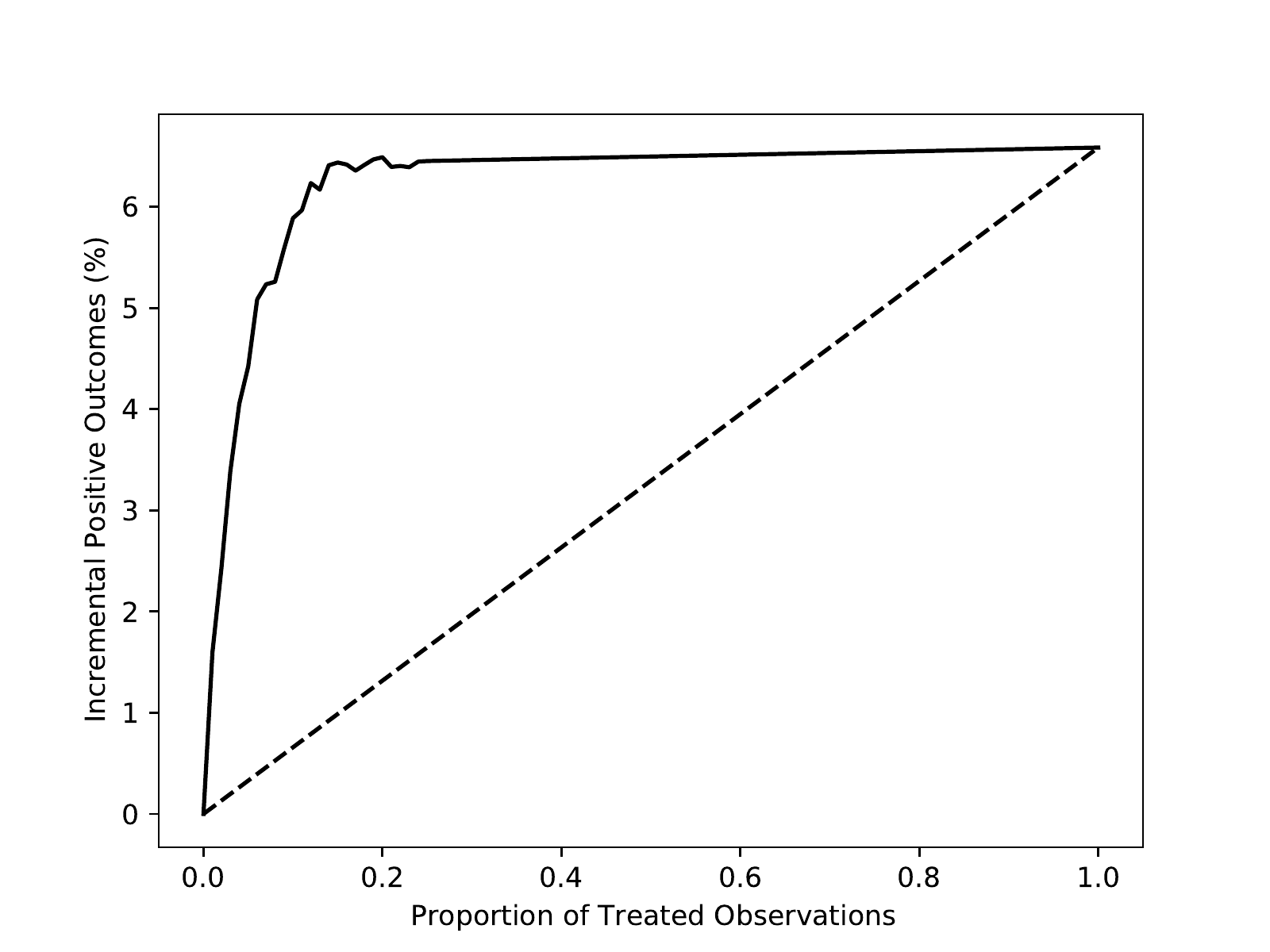}
    \includegraphics[width=0.49\textwidth]{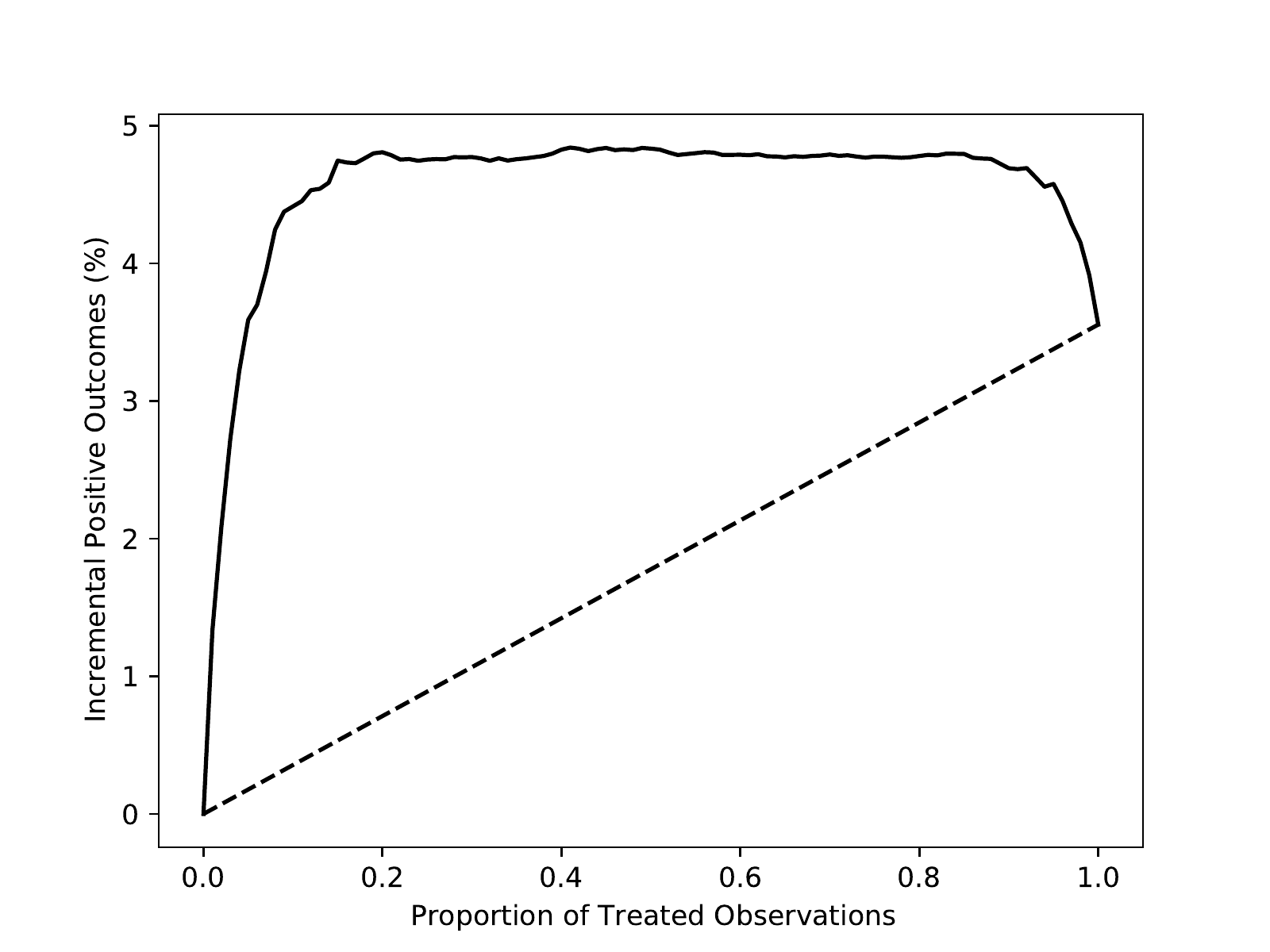}
    \includegraphics[width=0.49\textwidth]{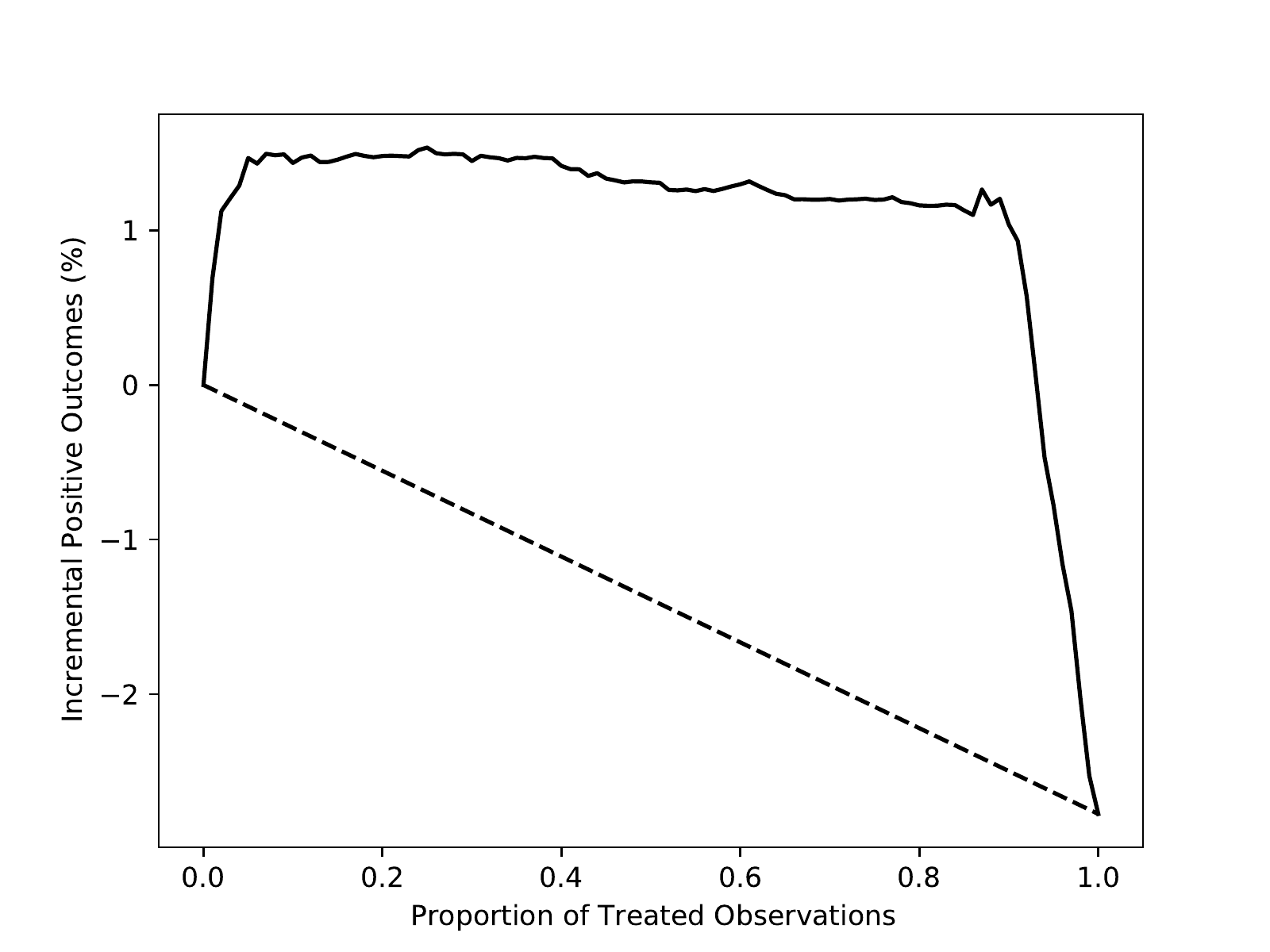}
    \includegraphics[width=0.49\textwidth]{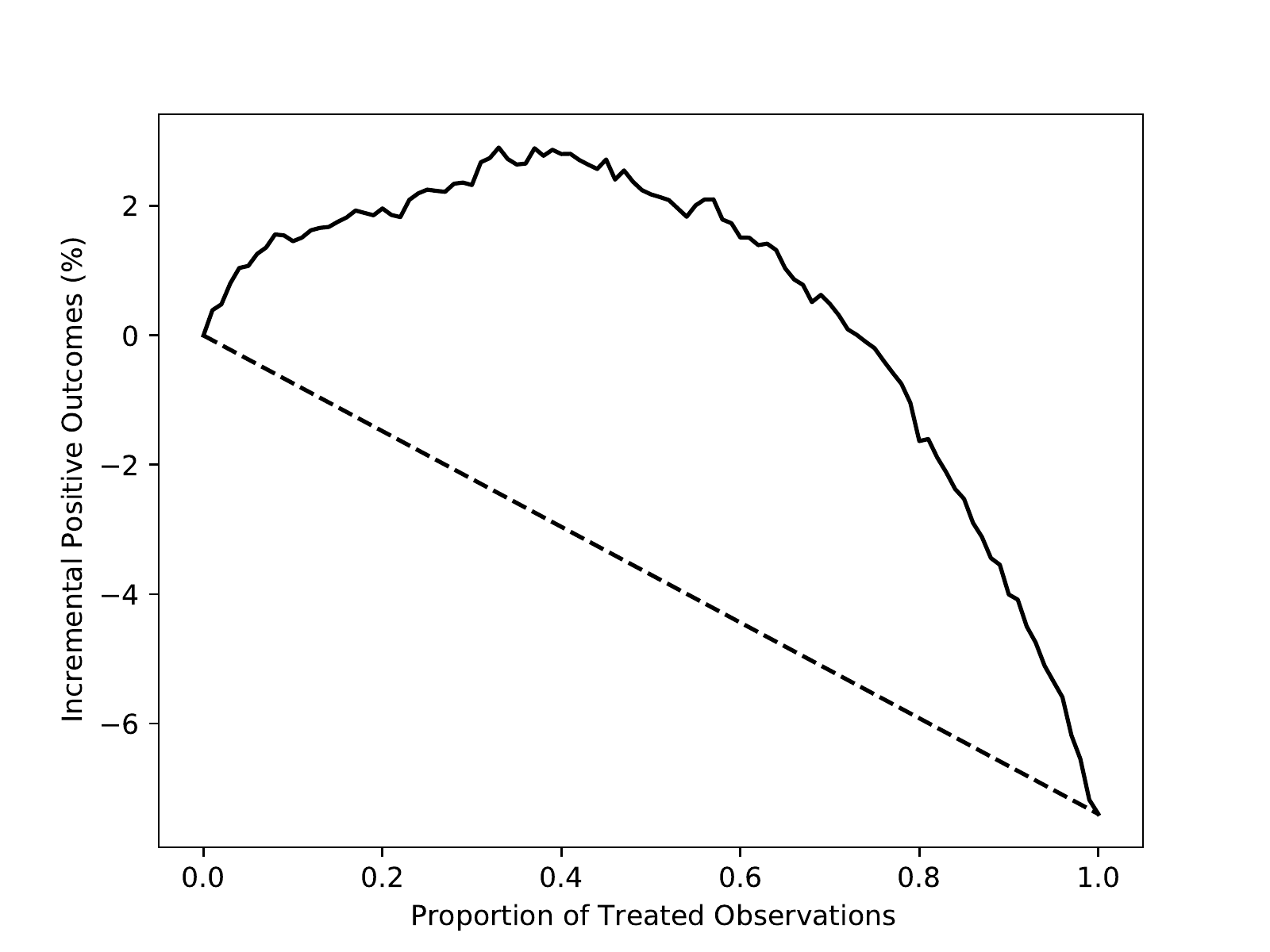}
    \caption{Qini curves based on the ``true" uplift with respect to scenarios 1-4 of Table~\ref{tab:scenarios}. Top left: scenario 1; top right: scenario 2; bottom left: scenario 3; bottom right: scenario 4.}
    \label{fig:qini_curves_scene}
\end{figure}

In Figure~\ref{fig:qini_curves_scene}, we show the Qini curves associated with the ``true" uplift $u^*$. The curve represents the incremental number of positive responses for a fraction of treated observations relative to the size of the treatment group in the sample. The dashed straight line is the performance of a random strategy. The scenarios are interesting because they represent different situations that we can face with real data. Indeed, scenario 1, although it seems simpler than the others (with a small noise level), is in fact a case that happens often, especially for marketing campaigns. In this scenario, by construction, there are no do-not-disturb clients. This results in an increasing monotonic Qini curve. In scenario 2, we introduce a group of do-not-disturb clients. In addition, we increase the noise level as well as the number of observations, but we reduce the number of variables. The average treatment effect is negative but a small group of persuadables is generated. In scenario 3, we introduce a quadratic treatment effect function and scenario 4 is the most complex one. \\

Each run of each simulation generates a dataset, which we split into training ($40\%$), validation ($30\%$), and test samples ($30\%$). We use the training and the validation sets to fit and fine-tune the models. Note that fine-tuning is specific to the type of the model, e.g., for a neural network, we fine-tune the learning rate and the regularization constants but for random forests, we fine-tune the number of trees as well as the depth of each tree. Then, we score the test samples with the fine-tuned models and compute the performance metric. We repeat each experiment $20$ times. For each simulated dataset, we implement the following models:
\begin{enumerate}
    \item[(a)] a multivariate logistic regression, i.e., a twin model with no hidden layer that optimizes the Binomial likelihood. This is the baseline model, and we will refer to it as \textit{Logistic}. For fair comparison, we use the twin neural architecture and $L_1$ regularization.
    \item[(b)] our twin models, i.e., models (with and without hidden layers) that optimize the uplift loss function \eqref{eq:augmented_loss_function} to estimate the regression parameters. We denote these models by \textit{Twin$_{\mu}$} and \textit{Twin$_{\mathrm{NN}}$}.
    \item[(c)] a Qini-based uplift regression model that uses several LHS structures to search for the optimal parameters (see \cite{belba2019qbased} for more details). We denote this model by \textit{Qini-based}.
    \item[(d)] two types of random forests designed for causal inference (see \cite{athey2019generalized} for more details). This method uses honest estimation, i.e., it does not use the same information for the partition of the covariates space and for the estimation of the uplift. We denote these models by \textit{Causal Forest} and \textit{Causal Forest (Honest)} without and with honest estimation respectively.
    \item[(e)] two types of random forests designed for uplift (see \cite{rzepakowski2010decision,guelman2012random} for more details). This method uses different split criteria. We denote these models by \textit{Uplift Random Forest (KL)} and \textit{Uplift Random Forest (ED)} for Kullback-Leibler and Euclidean split criterion respectively.
    \item[(f)] two types of R-learners (see \cite{NieWager2020Quasi} for more details). This method implements a residual-on-residual regression designed for CATE modeling. We consider two base models that we denote by {\it R-Learner }(\texttt{XGboost}) and {\it R-Learner }(\texttt{lasso}).
    \item[(g)] two types of X-learners (see \cite{kunzel2019metalearners} for more details). This method implements a three-stage regression designed for conditional treatment effect modeling. We consider two base models that we denote by {\it X-Learner }(\texttt{XGboost}) and {\it X-Learner }(\texttt{lasso}).
\end{enumerate}

\vspace{5mm}

All results are reported in Table~\ref{tab:all_scenarios}. If we pick “winners” in each of the simulation scenarios based on which method has the highest $\qadj$, {\it Twin$_{\mathrm{NN}}$} would win scenarios 1 and 4. In scenario 2, the {\it Qini-based}, {\it R-Learner }(\texttt{lasso}) and {\it X-Learner }(\texttt{lasso}) models yield a higher $\qadj$. In general the {\it Logistic} model is the worst. We observe that the random forests tend to overfit for several scenarios, with the exception of the causal forest with honest estimation which seems to mitigate this problem. In scenario 3, the {\it R-Learner }(\texttt{XGboost}) model has a slightly higher but not significant $\qadj$ than our 2-hidden layers {\it Twin$_{\mathrm{NN}}$} model.

\begin{table}[!ht]
    \centering
    \begin{tabular}{l|cccc}
        \hline
        Our implementation & \multicolumn{4}{c}{Scenarios}\\
        \hline
        Model (Size: total number of parameters) & 1 & 2 & 3 & 4  \\
        \hline
        {\it Logistic} ($2p+1$) & $0.75$ & $1.80$   & $0.98$ & $2.59$\\
        {\it Twin$_{\mu}$} ($2p+1$) & $1.04$  & $2.32$ & $1.20$ & $3.12$\\
        {\it Twin$_{\mathrm{NN}}$} ($(p+1)\times 64 + 64$)  & $1.54$  & $2.61$  & $1.22$ & $3.10$ \\
        {\it Twin$_{\mathrm{NN}}$} ($(p+1)\times 128 + 128$) & $1.56$  & $2.54$  & $1.28$ & $3.16$\\
        {\it Twin$_{\mathrm{NN}}$} ($(p+1)\times (2p+1) + (2p+1)$) & $ 1.63$ & $2.65$  & $1.30$ & $3.14$\\
        {\it Twin$_{\mathrm{NN}}$} ($(p+1)\times 256 + 256$) & $1.57$ & $2.65$  & $ 1.32$ & $3.26$ \\
        {\it Twin$_{\mathrm{NN}}$} ($(p+1)\times 512 + 512$) & $1.60$ & $2.49$  & $1.18$ & $ 3.35$\\
        {\it Twin$_{\mathrm{NN}}$} with structured sparsity ($(p+1)\times 512 + 512$) & $\bf 1.67$  & $\bf 2.67$ & $1.35$ & $\bf 3.58$\\
        2-hidden layers {\it Twin$_{\mathrm{NN}}$} ($(p+1)\times(p+1) + (p+1)\times(p) + p $) & $1.59$ & $2.36$  & $\bf 1.38$ & $3.39$ \\
        \hline
        \multicolumn{5}{l}{Open-source implementation}\\
        \hline
        {\it Qini-based} \citep{belba2019qbased} & $\bf 1.02$ & $2.68$  & $1.09$ & $2.94$\\
        {\it Causal Forest} \citep{athey2019generalized} & $0.75$ & $2.22$  & $0.94$ & $2.79$ \\
        {\it Causal Forest (Honest)} \citep{athey2019generalized} & $0.75$  & $2.51$  & $1.14$ & $\bf 3.07$\\
        {\it Uplift Random Forest (KL)} \citep{guelman2012random} & $0.74$ & $2.52$  & $1.01$& $2.19$ \\
        {\it Uplift Random Forest (ED)} \citep{guelman2012random} & $0.68$ & $2.42$ & $0.99$ & $2.33$ \\
        {\it R-Learner }(\texttt{XGboost}) \citep{NieWager2020Quasi} & $0.76$ & $2.63$ & $\bf 1.40$ & $2.12$\\
        {\it R-Learner }(\texttt{lasso}) \citep{NieWager2020Quasi} & $0.66$ & $2.75$ & $0.87$ & $2.83$ \\
        {\it X-Learner }(\texttt{XGboost}) \citep{kunzel2019metalearners} & $0.72$  & $2.57$ & $1.31$ & $2.37$ \\
        {\it X-Learner }(\texttt{lasso}) \citep{kunzel2019metalearners} & $0.77$ & $\bf 2.78$ & $0.77$ & $2.91$ \\
        \hline
    \end{tabular}
    \vspace{2mm}
    \caption{Summary: models comparison in terms of $\qadj$ averaged on the test set over $20$ runs. Note that the maximum standard-error is $0.15$; we do not report them to simplify the Table. The model size indicates the number of parameters to estimate (excluding the bias terms).}
    \label{tab:all_scenarios}
\end{table}

Now, looking at our 1-hidden layer {\it Twin$_{\mathrm{NN}}$} with fixed number of hidden neurons, for scenario 1, in which $p = 200$, we see that the best model is reached for $m = 512$; for scenario 2, the model performs best with $m = 201$ or $m = 256$ and with $m = 256$ and $m = 512$ for scenarios 3 and 4. 
As discussed earlier, the use of the scaling factor penalization on the weight matrix allows to automatically fine-tune the number of hidden neurons (i.e., pruning). This has the effect of increasing the performance of the underlying models from a predictive point of view, as shown in Table~\ref{tab:all_scenarios}. 
Finally, 
in our experiments, increasing the number of hidden layers did not improve the performance significantly. When we fix the initial number of hidden neurons to $p+1$ and $p$ for the $2$ hidden layer, in scenarios 1 and 4, the average $\qadj$ are $1.59$ and $3.39$ respectively.  

\section*{Computational details}

In all of our experiments, we used available implementations of the benchmarks. We used the \proglang{R4.0.3} libraries \pkg{grf} \citep{tibshirani2019R_grf} and \pkg{uplift} \citep{guelman2014R_uplift} for the random forests implementations. We also used the \proglang{R4.0.3} library \pkg{tools4uplift} \citep{belba2020R_tools4uplift} to fit the Qini-based uplift regression. Finally, we used the \proglang{R4.0.3} library \pkg{rlearner} \citep{nie2020rlearner} to fit the {\it R-} and {\it X-learners}. Note that the {\it Logistic} and \textit{Twin}  methods were implemented with \pkg{Pytorch1.3.1} in \proglang{Python3.7.5}. 

\section*{Acknowledgements}
Mouloud Belbahri and Alejandro Murua were partially funded by The Natural Sciences and Engineering Research Council of Canada grant~2019-05444. Vahid Partovi Nia was supported by the Natural Sciences and Engineering Research Council of Canada grant~418034-2012. Mouloud Belbahri wants to acknowledge \proglang{Python} implementation discussions with Ghaith Kazma and Eyy{\"u}b Sari. Mouloud Belbahri and Olivier Gandouet thank Julie Hussin for proof reading an early version of the manuscript and for interesting discussions about potential applications in Bioinformatics.

\bibliographystyle{plainnat}
\bibliography{mybib}

\begin{thebibliography}{61}
\providecommand{\natexlab}[1]{#1}
\providecommand{\url}[1]{\texttt{#1}}
\expandafter\ifx\csname urlstyle\endcsname\relax
  \providecommand{\doi}[1]{doi: #1}\else
  \providecommand{\doi}{doi: \begingroup \urlstyle{rm}\Url}\fi

\bibitem[Alemi et~al.(2009)Alemi, Erdman, Griva, and
  Evans]{Alemi.etal-PersonalizedMedicine-2009}
Farrokh Alemi, Harold Erdman, Igor Griva, and Charles~H Evans.
\newblock Improved statistical methods are needed to advance personalized
  medicine.
\newblock \emph{The Open Translational Medicine Journal}, 1:\penalty0 16, 2009.

\bibitem[Arlot et~al.(2010)Arlot, Celisse, et~al.]{arlot2010survey}
Sylvain Arlot, Alain Celisse, et~al.
\newblock A survey of cross-validation procedures for model selection.
\newblock \emph{Statistics Surveys}, 4:\penalty0 40--79, 2010.

\bibitem[Athey and Imbens(2015)]{athey2015machine}
Susan Athey and Guido~W Imbens.
\newblock Machine learning methods for estimating heterogeneous causal effects.
\newblock \emph{stat}, 1050\penalty0 (5):\penalty0 1--26, 2015.

\bibitem[Athey et~al.(2019)Athey, Tibshirani, Wager,
  et~al.]{athey2019generalized}
Susan Athey, Julie Tibshirani, Stefan Wager, et~al.
\newblock Generalized random forests.
\newblock \emph{The Annals of Statistics}, 47\penalty0 (2):\penalty0
  1148--1178, 2019.

\bibitem[Beck(2017)]{beck2017first}
Amir Beck.
\newblock \emph{First-order methods in optimization}.
\newblock SIAM, 2017.

\bibitem[Belbahri et~al.(2019)Belbahri, Murua, Gandouet, and
  Partovi~Nia]{belba2019qbased}
Mouloud Belbahri, Alejandro Murua, Olivier Gandouet, and Vahid Partovi~Nia.
\newblock Qini-based uplift regression.
\newblock \emph{arXiv preprint arXiv:1911.12474}, 2019.

\bibitem[Belbahri et~al.(2020)Belbahri, Gandouet, Murua, and
  Nia]{belba2020R_tools4uplift}
Mouloud Belbahri, Olivier Gandouet, Alejandro Murua, and Vahid~Partovi Nia.
\newblock \emph{tools4uplift: Tools for Uplift Modeling}, 2020.
\newblock URL \url{https://CRAN.R-project.org/package=tools4uplift}.
\newblock R package version 1.0.0.

\bibitem[Bottou et~al.(2018)Bottou, Curtis, and
  Nocedal]{bottou2018optimization}
L{\'e}on Bottou, Frank~E Curtis, and Jorge Nocedal.
\newblock Optimization methods for large-scale machine learning.
\newblock \emph{SIAM Review}, 60\penalty0 (2):\penalty0 223--311, 2018.

\bibitem[Breiman(2001)]{Breiman-RandomForest-2001}
Leo Breiman.
\newblock Random forests.
\newblock \emph{Machine Learning}, 45\penalty0 (1):\penalty0 5--32, 2001.

\bibitem[Breiman et~al.(1984)Breiman, Friedman, Stone, and
  Olshen]{breiman1984classification}
Leo Breiman, Jerome Friedman, Charles~J Stone, and Richard~A Olshen.
\newblock \emph{Classification and Regression Trees}.
\newblock CRC press, 1984.

\bibitem[Bromley et~al.(1994)Bromley, Guyon, LeCun, S{\"a}ckinger, and
  Shah]{bromley1994signature}
Jane Bromley, Isabelle Guyon, Yann LeCun, Eduard S{\"a}ckinger, and Roopak
  Shah.
\newblock Signature verification using a ``siamese" time delay neural network.
\newblock In \emph{Advances in Neural Information Processing Systems}, pages
  737--744, 1994.

\bibitem[Chen et~al.(2015)Chen, He, Benesty, Khotilovich, Tang, Cho,
  et~al.]{chen2015xgboost}
Tianqi Chen, Tong He, Michael Benesty, Vadim Khotilovich, Yuan Tang, Hyunsu
  Cho, et~al.
\newblock Xgboost: extreme gradient boosting.
\newblock \emph{R package version 0.4-2}, 1\penalty0 (4), 2015.

\bibitem[Chipman et~al.(2010)Chipman, George, McCulloch,
  et~al.]{chipman2010bart}
Hugh~A Chipman, Edward~I George, Robert~E McCulloch, et~al.
\newblock Bart: Bayesian additive regression trees.
\newblock \emph{The Annals of Applied Statistics}, 4\penalty0 (1):\penalty0
  266--298, 2010.

\bibitem[Chopra et~al.(2005)Chopra, Hadsell, and LeCun]{chopra2005learning}
Sumit Chopra, Raia Hadsell, and Yann LeCun.
\newblock Learning a similarity metric discriminatively, with application to
  face verification.
\newblock In \emph{2005 IEEE Computer Society Conference on Computer Vision and
  Pattern Recognition (CVPR'05)}, volume~1, pages 539--546. IEEE, 2005.

\bibitem[Cover and Hart(1967)]{cover1967nearest}
Thomas Cover and Peter Hart.
\newblock Nearest neighbor pattern classification.
\newblock \emph{IEEE transactions on information theory}, 13\penalty0
  (1):\penalty0 21--27, 1967.

\bibitem[Crump et~al.(2008)Crump, Hotz, Imbens, and
  Mitnik]{crump2008nonparametric}
Richard~K Crump, V~Joseph Hotz, Guido~W Imbens, and Oscar~A Mitnik.
\newblock Nonparametric tests for treatment effect heterogeneity.
\newblock \emph{The Review of Economics and Statistics}, 90\penalty0
  (3):\penalty0 389--405, 2008.

\bibitem[Cybenko(1989)]{cybenko1989approximation}
George Cybenko.
\newblock Approximation by superpositions of a sigmoidal function.
\newblock \emph{Mathematics of Control, Signals and Systems}, 2\penalty0
  (4):\penalty0 303--314, 1989.

\bibitem[{Diemert Eustache, Betlei Artem} et~al.(2018){Diemert Eustache, Betlei
  Artem}, Renaudin, and Massih-Reza]{Diemert2018}
{Diemert Eustache, Betlei Artem}, Christophe Renaudin, and Amini Massih-Reza.
\newblock A large scale benchmark for uplift modeling.
\newblock In \emph{Proceedings of the AdKDD and TargetAd Workshop, KDD,
  London,United Kingdom, August, 20, 2018}. ACM, 2018.

\bibitem[Efron and Hastie(2016)]{efron2016computer}
Bradley Efron and Trevor Hastie.
\newblock \emph{Computer age statistical inference}, volume~5.
\newblock Cambridge University Press, 2016.

\bibitem[Friedman et~al.(2007)Friedman, Hastie, H{\"o}fling, Tibshirani,
  et~al.]{friedman2007pathwise}
Jerome Friedman, Trevor Hastie, Holger H{\"o}fling, Robert Tibshirani, et~al.
\newblock Pathwise coordinate optimization.
\newblock \emph{The Annals of Applied Statistics}, 1\penalty0 (2):\penalty0
  302--332, 2007.

\bibitem[Guelman(2014)]{guelman2014R_uplift}
Leo Guelman.
\newblock \emph{uplift: Uplift Modeling}, 2014.
\newblock URL \url{https://CRAN.R-project.org/package=uplift}.
\newblock R package version 0.3.5.

\bibitem[Guelman et~al.(2012)Guelman, Guill{\'e}n, and
  P{\'e}rez-Mar{\'\i}n]{guelman2012random}
Leo Guelman, Montserrat Guill{\'e}n, and Ana~M P{\'e}rez-Mar{\'\i}n.
\newblock Random forests for uplift modeling: an insurance customer retention
  case.
\newblock In \emph{Modeling and Simulation in Engineering, Economics and
  Management}, pages 123--133. Springer, 2012.

\bibitem[Hansotia and Rukstales(2001)]{hansotia2002direct}
Behram Hansotia and Bradley Rukstales.
\newblock Direct marketing for multichannel retailers: Issues, challenges and
  solutions.
\newblock \emph{Journal of Database Marketing and Customer Strategy
  Management}, 9\penalty0 (3):\penalty0 259--266, 2001.

\bibitem[Holland(1986)]{holland1986statistics}
Paul~W Holland.
\newblock Statistics and causal inference.
\newblock \emph{Journal of the American Statistical Association}, 81\penalty0
  (396):\penalty0 945--960, 1986.

\bibitem[Hornik(1991)]{hornik1991approximation}
Kurt Hornik.
\newblock Approximation capabilities of multilayer feedforward networks.
\newblock \emph{Neural Networks}, 4\penalty0 (2):\penalty0 251--257, 1991.

\bibitem[Jaskowski and Jaroszewicz(2012)]{jaskowski2012uplift}
Maciej Jaskowski and Szymon Jaroszewicz.
\newblock Uplift modeling for clinical trial data.
\newblock In \emph{ICML Workshop on Clinical Data Analysis}, 2012.

\bibitem[Karpathy et~al.(2014)Karpathy, Toderici, Shetty, Leung, Sukthankar,
  and Fei-Fei]{karpathy2014large}
Andrej Karpathy, George Toderici, Sanketh Shetty, Thomas Leung, Rahul
  Sukthankar, and Li~Fei-Fei.
\newblock Large-scale video classification with convolutional neural networks.
\newblock In \emph{Proceedings of the IEEE conference on Computer Vision and
  Pattern Recognition}, pages 1725--1732, 2014.

\bibitem[K{\"u}nzel et~al.(2019)K{\"u}nzel, Sekhon, Bickel, and
  Yu]{kunzel2019metalearners}
S{\"o}ren~R K{\"u}nzel, Jasjeet~S Sekhon, Peter~J Bickel, and Bin Yu.
\newblock Metalearners for estimating heterogeneous treatment effects using
  machine learning.
\newblock \emph{Proceedings of the National Academy of Sciences}, 116\penalty0
  (10):\penalty0 4156--4165, 2019.

\bibitem[Lamont et~al.(2018)Lamont, Lyons, Jaki, Stuart, Feaster, Tharmaratnam,
  Oberski, Ishwaran, Wilson, and Van~Horn]{lamont2018identification}
Andrea Lamont, Michael~D Lyons, Thomas Jaki, Elizabeth Stuart, Daniel~J
  Feaster, Kukatharmini Tharmaratnam, Daniel Oberski, Hemant Ishwaran, Dawn~K
  Wilson, and M~Lee Van~Horn.
\newblock Identification of predicted individual treatment effects in
  randomized clinical trials.
\newblock \emph{Statistical Methods in Medical Research}, 27\penalty0
  (1):\penalty0 142--157, 2018.

\bibitem[Lin et~al.(2015)Lin, Cui, Belongie, and Hays]{lin2015learning}
Tsung-Yi Lin, Yin Cui, Serge Belongie, and James Hays.
\newblock Learning deep representations for ground-to-aerial geolocalization.
\newblock In \emph{Proceedings of the IEEE conference on Computer Vision and
  Pattern Recognition}, pages 5007--5015, 2015.

\bibitem[Lo(2002)]{lo2002true}
Victor~SY Lo.
\newblock The true lift model: a novel data mining approach to response
  modeling in database marketing.
\newblock \emph{ACM SIGKDD Explorations Newsletter}, 4\penalty0 (2):\penalty0
  78--86, 2002.

\bibitem[McKay et~al.(2000)McKay, Beckman, and Conover]{mckay2000comparison}
Michael~D McKay, Richard~J Beckman, and William~J Conover.
\newblock A comparison of three methods for selecting values of input variables
  in the analysis of output from a computer code.
\newblock \emph{Technometrics}, 42\penalty0 (1):\penalty0 55--61, 2000.

\bibitem[Moher et~al.(2012)Moher, Hopewell, Schulz, Montori, G{\o}tzsche,
  Devereaux, Elbourne, Egger, and Altman]{moher2012consort}
David Moher, Sally Hopewell, Kenneth~F Schulz, Victor Montori, Peter~C
  G{\o}tzsche, PJ~Devereaux, Diana Elbourne, Matthias Egger, and Douglas~G
  Altman.
\newblock Consort 2010 explanation and elaboration: updated guidelines for
  reporting parallel group randomised trials.
\newblock \emph{International Journal of Surgery}, 10\penalty0 (1):\penalty0
  28--55, 2012.

\bibitem[Mosci et~al.(2010)Mosci, Rosasco, Santoro, Verri, and
  Villa]{mosci2010solving}
Sofia Mosci, Lorenzo Rosasco, Matteo Santoro, Alessandro Verri, and Silvia
  Villa.
\newblock Solving structured sparsity regularization with proximal methods.
\newblock In \emph{Joint European conference on Machine Learning and Knowledge
  Discovery in Databases}, pages 418--433. Springer, 2010.

\bibitem[Neyman(1923)]{Neyman:1923}
Jerzy~S Neyman.
\newblock On the application of probability theory to agricultural experiments.
\newblock \emph{Annals of Agricultural Sciences}, 10:\penalty0 1--51, 1923.

\bibitem[Nie and Wager(2020)]{NieWager2020Quasi}
Xinkun Nie and Stefan Wager.
\newblock Quasi-oracle estimation of heterogeneous treatment effects.
\newblock \emph{Biometrika}, 2020.

\bibitem[Nie et~al.(2020)Nie, Schuler, and Wager]{nie2020rlearner}
Xinkun Nie, Alejandro Schuler, and Stefan Wager.
\newblock \emph{rlearner: R-learner for Heterogeneous Treatment Effect
  Estimation}, 2020.
\newblock R package version 1.1.0.

\bibitem[Parkhi et~al.(2015)Parkhi, Vedaldi, and Zisserman]{parkhi2015deep}
Omkar~M Parkhi, Andrea Vedaldi, and Andrew Zisserman.
\newblock Deep face recognition.
\newblock \emph{British Machine Vision Association}, 2015.

\bibitem[Pinkus(1999)]{pinkus1999approximation}
Allan Pinkus.
\newblock Approximation theory of the {MLP} model in neural networks.
\newblock \emph{Acta Numerica}, 8\penalty0 (1):\penalty0 143--195, 1999.

\bibitem[Powers et~al.(2018)Powers, Qian, Jung, Schuler, Shah, Hastie, and
  Tibshirani]{powers2018some}
Scott Powers, Junyang Qian, Kenneth Jung, Alejandro Schuler, Nigam~H Shah,
  Trevor Hastie, and Robert Tibshirani.
\newblock Some methods for heterogeneous treatment effect estimation in high
  dimensions.
\newblock \emph{Statistics in Medicine}, 37\penalty0 (11):\penalty0 1767--1787,
  2018.

\bibitem[Radcliffe and Surry(2011)]{radcliffe2011real}
Nicholas~J Radcliffe and Patrick~D Surry.
\newblock Real-world uplift modelling with significance-based uplift trees.
\newblock \emph{White Paper TR-2011-1, Stochastic Solutions}, 2011.

\bibitem[Radcliffe(2007)]{radcliffe2007using}
NJ~Radcliffe.
\newblock Using control groups to target on predicted lift: Building and
  assessing uplift models.
\newblock \emph{Direct Market J Direct Market Assoc Anal Council}, 1:\penalty0
  14--21, 2007.

\bibitem[Radcliffe and Surry(1999)]{radcliffe1999differential}
NJ~Radcliffe and PD~Surry.
\newblock Differential response analysis: Modeling true response by isolating
  the effect of a single action.
\newblock \emph{Credit Scoring and Credit Control VI. Edinburgh, Scotland},
  1999.

\bibitem[Ramakrishnan et~al.(2020)Ramakrishnan, Sari, and
  Nia]{ramakrishnan2020differentiable}
Ramchalam~Kinattinkara Ramakrishnan, Eyyub Sari, and Vahid~Partovi Nia.
\newblock Differentiable mask for pruning convolutional and recurrent networks.
\newblock In \emph{2020 17th Conference on Computer and Robot Vision (CRV)},
  pages 222--229. IEEE, 2020.

\bibitem[Robbins and Monro(1951)]{robbins1951stochastic}
Herbert Robbins and Sutton Monro.
\newblock A stochastic approximation method.
\newblock \emph{The Annals of Mathematical Statistics}, pages 400--407, 1951.

\bibitem[Robinson(1988)]{robinson1988root}
Peter~M Robinson.
\newblock Root-n-consistent semiparametric regression.
\newblock \emph{Econometrica: Journal of the Econometric Society}, pages
  931--954, 1988.

\bibitem[Rosenbaum and Rubin(1983)]{rosenbaum1983central}
Paul~R Rosenbaum and Donald~B Rubin.
\newblock The central role of the propensity score in observational studies for
  causal effects.
\newblock \emph{Biometrika}, 70\penalty0 (1):\penalty0 41--55, 1983.

\bibitem[Rubin(1974)]{rubin1974estimating}
Donald~B Rubin.
\newblock Estimating causal effects of treatments in randomized and
  nonrandomized studies.
\newblock \emph{Journal of Educational Psychology}, 66\penalty0 (5):\penalty0
  688, 1974.

\bibitem[Rzepakowski and Jaroszewicz(2010)]{rzepakowski2010decision}
Piotr Rzepakowski and Szymon Jaroszewicz.
\newblock Decision trees for uplift modeling.
\newblock In \emph{2010 IEEE International Conference on Data Mining}, pages
  441--450. IEEE, 2010.

\bibitem[Schroff et~al.(2015)Schroff, Kalenichenko, and
  Philbin]{schroff2015facenet}
Florian Schroff, Dmitry Kalenichenko, and James Philbin.
\newblock Facenet: A unified embedding for face recognition and clustering.
\newblock In \emph{Proceedings of the IEEE conference on Computer Vision and
  Pattern Recognition}, pages 815--823, 2015.

\bibitem[So{\l}tys et~al.(2015)So{\l}tys, Jaroszewicz, and
  Rzepakowski]{soltys2015ensemble}
Micha{\l} So{\l}tys, Szymon Jaroszewicz, and Piotr Rzepakowski.
\newblock Ensemble methods for uplift modeling.
\newblock \emph{Data Mining and Knowledge Discovery}, 29\penalty0 (6):\penalty0
  1531--1559, 2015.

\bibitem[Su et~al.(2009)Su, Tsai, Wang, Nickerson, and Li]{su2009subgroup}
Xiaogang Su, Chih-Ling Tsai, Hansheng Wang, David~M Nickerson, and Bogong Li.
\newblock Subgroup analysis via recursive partitioning.
\newblock \emph{Journal of Machine Learning Research}, 10\penalty0
  (Feb):\penalty0 141--158, 2009.

\bibitem[Su et~al.(2012)Su, Kang, Fan, Levine, and Yan]{su2012facilitating}
Xiaogang Su, Joseph Kang, Juanjuan Fan, Richard~A Levine, and Xin Yan.
\newblock Facilitating score and causal inference trees for large observational
  studies.
\newblock \emph{Journal of Machine Learning Research}, 13\penalty0
  (Oct):\penalty0 2955--2994, 2012.

\bibitem[Taigman et~al.(2014)Taigman, Yang, Ranzato, and
  Wolf]{taigman2014deepface}
Yaniv Taigman, Ming Yang, Marc'Aurelio Ranzato, and Lior Wolf.
\newblock Deepface: Closing the gap to human-level performance in face
  verification.
\newblock In \emph{Proceedings of the IEEE conference on Computer Vision and
  Pattern Recognition}, pages 1701--1708, 2014.

\bibitem[Tibshirani et~al.(2019)Tibshirani, Athey, and
  Wager]{tibshirani2019R_grf}
Julie Tibshirani, Susan Athey, and Stefan Wager.
\newblock \emph{grf: Generalized Random Forests}, 2019.
\newblock URL \url{https://CRAN.R-project.org/package=grf}.
\newblock R package version 0.10.4.

\bibitem[Tibshirani(1996)]{Tibshirani_lasso_1996}
R.~Tibshirani.
\newblock Regression shrinkage and selection via the lasso.
\newblock \emph{Journal of the Royal Statistical Society. Series B
  (Methodological)}, pages 267--288, 1996.

\bibitem[Tsang et~al.(2018)Tsang, Liu, Purushotham, Murali, and
  Liu]{tsang2018neural}
Michael Tsang, Hanpeng Liu, Sanjay Purushotham, Pavankumar Murali, and Yan Liu.
\newblock Neural interaction transparency (nit): Disentangling learned
  interactions for improved interpretability.
\newblock In \emph{Advances in Neural Information Processing Systems}, pages
  5804--5813, 2018.

\bibitem[Wager and Athey(2018)]{wager2018estimation}
Stefan Wager and Susan Athey.
\newblock Estimation and inference of heterogeneous treatment effects using
  random forests.
\newblock \emph{Journal of the American Statistical Association}, 113\penalty0
  (523):\penalty0 1228--1242, 2018.

\bibitem[Wilcoxon(1992)]{wilcoxon1992individual}
Frank Wilcoxon.
\newblock Individual comparisons by ranking methods.
\newblock In \emph{Breakthroughs in Statistics}, pages 196--202. Springer,
  1992.

\bibitem[Yuan and Lin(2006)]{yuan2006model}
Ming Yuan and Yi~Lin.
\newblock Model selection and estimation in regression with grouped variables.
\newblock \emph{Journal of the Royal Statistical Society: Series B (Statistical
  Methodology)}, 68\penalty0 (1):\penalty0 49--67, 2006.

\bibitem[Zhao et~al.(2017)Zhao, Fang, and Simchi-Levi]{zhao2017practically}
Yan Zhao, Xiao Fang, and David Simchi-Levi.
\newblock A practically competitive and provably consistent algorithm for
  uplift modeling.
\newblock In \emph{2017 IEEE International Conference on Data Mining (ICDM)},
  pages 1171--1176. IEEE, 2017.

\end{thebibliography}

\end{document}